%% file: main.tex
\documentclass{article} % For LaTeX2e
\usepackage{iclr2027_conference,times}

\input{math_commands.tex}

\usepackage{hyperref}
\usepackage{url}
\usepackage{booktabs}
\usepackage{graphicx}
\usepackage{threeparttable}

\usepackage{graphicx}
\usepackage{wrapfig}

\usepackage{amsmath}
\usepackage{amssymb}
\usepackage{mathtools}
\usepackage{algorithm}
\usepackage{algpseudocode}
\usepackage{makecell}
\usepackage{tabularx,xcolor}
\usepackage{caption}

\usepackage{ragged2e}
\usepackage{fvextra}
\fvset{breaksymbolleft={}}

\usepackage[most]{tcolorbox}

\newtcblisting{DigestPromptBox}[1]{
  listing only,
  breakable,
  title={\footnotesize #1},
  top=1mm,
  bottom=1mm,
  listing options={
    basicstyle=\fontencoding{T1}\selectfont\ttfamily\scriptsize,
    breaklines=true,
    breakatwhitespace=false,
    columns=fullflexible,
    keepspaces=true,
    showstringspaces=false
  }
}

\title{DIGEST: Discovery-Guided LLM Tuning for Symbolic Regression}

\author{%
Kaifeng Jin$^1$, Ruizhong Qiu$^1$, Zihao Li$^1$, Xiao Lin$^1$, Dongqi Fu$^2$\\
\textbf{Yinglong Xia$^2$, Hong Li$^2$, Hanghang Tong$^1$}\\
$^1$University of Illinois Urbana-Champaign, $^2$Meta
}

\newcommand{\methodname}{DIGEST}

\iclrpreprint
\begin{document}

\maketitle

\input{sections/0-abs}
\input{sections/1-intro}
\input{sections/2-prelim}
\input{sections/3-method}

\input{sections/4-exp}
\input{sections/5-ablation}
\input{sections/6-related}
\input{sections/7-con}

\subsection*{AI use statement}
We used generative AI tools to assist with editing and polishing the manuscript. All AI-assisted work was reviewed by the authors. The authors take full responsibility for the final
content and claims of this work.

\subsection*{Reproducibility statement}

Section 3, Section 4 and Algorithm~1 describe the proposed method and experimental
setup. Appendix~A provides detailed implementation and configuration
information, including the symbolic interface, online objective, GP and budget
settings, and adaptation controls. Appendix~B specifies the benchmark
definitions and evaluation protocols, while Appendix~C documents baseline
configurations and timing details. Appendix~D provides detailed experimental
results, and Appendix~E reproduces the active prompt templates used for the
LLM-based baselines. We plan to make the implementation publicly available
upon acceptance.

\bibliography{reference}
\bibliographystyle{iclr2027_conference}

\appendix
\input{sections/app}

\end{document}

%% file: math_commands.tex
\usepackage{amsmath,amsfonts,bm}

\def\eqref#1{equation~\ref{#1}}
\def\1{\bm{1}}

\DeclareMathAlphabet{\mathsfit}{\encodingdefault}{\sfdefault}{m}{sl}
\SetMathAlphabet{\mathsfit}{bold}{\encodingdefault}{\sfdefault}{bx}{n}

%% file: sections/0-abs.tex
\begin{abstract}
Symbolic regression (SR) seeks to discover mathematical expressions directly from data, without requiring task-specific scientific descriptions. Despite tremendous progress in both conventional search-based methods and recent LLM-based methods, two critical limitations largely remain. First (\textit{Limited Generation Flexibility}), conventional search-based methods rely on either fixed search rules or specialized learned generators, severely limiting adaptive candidate generation across diverse expression structures. Second (\textit{Underutilized Search Discoveries}), most LLM-based methods cannot reliably recover ground-truth expressions because they fall short in effectively leveraging the LLM itself to thoroughly digest past expressions or adaptively guide future search. To address these limitations, we propose \methodname. To render greater flexibility in generation, \methodname\ repurposes a pretrained decoder block from a lightweight LLM into a symbolic policy for Genetic Programming (GP) population seeding, thereby leveraging the broader modeling capacity of the LLM. To more thoroughly utilize search discoveries, high-quality expressions discovered by GP are fed back to update the LLM-derived policy online, thereby enabling the LLM to progressively internalize past search discoveries and translate them into improved subsequent population generation. Experiments show that on two ground-truth benchmarks, \methodname\ outperforms the strongest competitor by up to $2.59\%$ in recovery. Among LLM-based baselines, it improves recovery by at least $54.91\%$ while using at most $17.4\%$ of the strongest competitor's mean search time. On 122 BlackBox datasets, \methodname\ achieves the highest accuracy while maintaining a moderate model complexity.
\end{abstract}

%% file: sections/1-intro.tex
\section{Introduction}
\label{sec:introduction}

\begin{wrapfigure}{r}{0.48\textwidth}
    \centering
    \includegraphics[width=\linewidth]
    {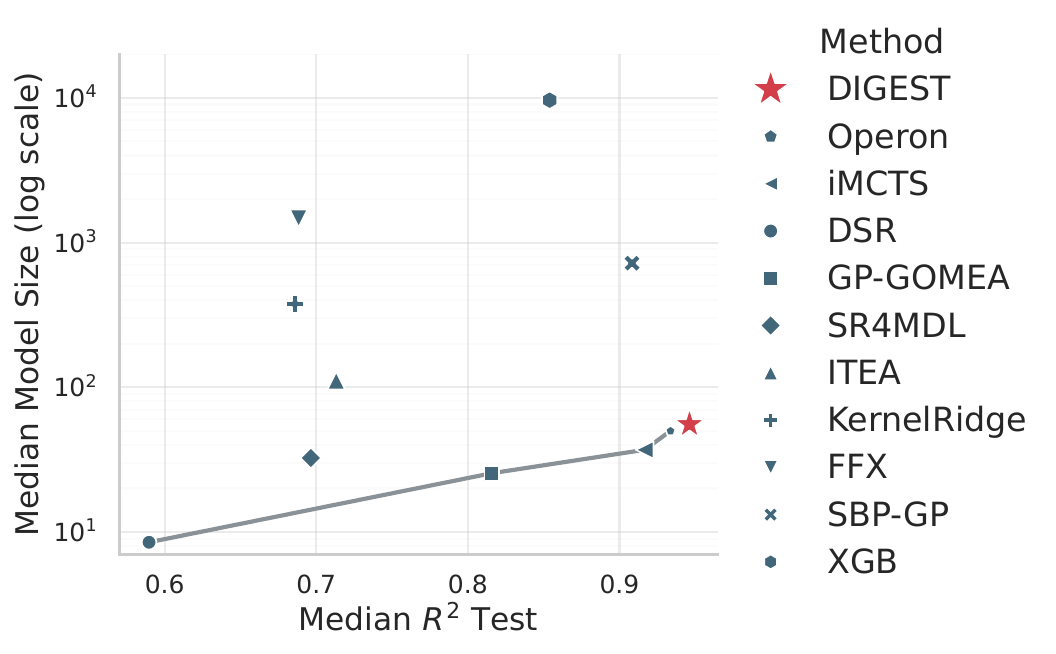}
    \caption{\methodname\ achieves the highest $R^2$ with moderate model size on BlackBox datasets.}
    \label{fig:blackbox_teaser}
\end{wrapfigure}

Symbolic regression (SR) seeks to discover mathematical expressions directly
from data, without requiring task-specific scientific descriptions
\citep{la2021contemporary}.
By searching over both expression structure and numerical parameters, SR
provides a way to construct explicit models when a suitable functional form
is not known in advance. Recent large language model (LLM) approaches further use scientific
descriptions and domain knowledge as auxiliary information for equation
discovery \citep{shojaee2025llm}. Such information can be valuable when available. However, there are also black-box discovery settings in which only numerical
input-output observations are available, without reliable task-specific descriptions or variable semantics. In such settings, SR is particularly valuable because candidate functional forms cannot be readily specified from domain knowledge and must instead be
discovered from the observations themselves. Accordingly, SR methods, particularly LLM-based ones, should remain effective
even when scientific context is unavailable, rather than relying on it as a prerequisite.
We focus on this classical data-driven SR setting, using input-output
observations and a predefined symbolic vocabulary without supplied variable
semantics or physical units. 

Despite tremendous progress in both conventional search-based and recent LLM-based SR, two critical limitations remain. First (\emph{Limited Generation Flexibility}), conventional search-based methods generate candidates through predefined
search mechanisms or specialized learned generators. For example, PySR evolves
expression populations through genetic operations
\citep{cranmer2023interpretable}, DSR learns a recurrent generation policy
\citep{petersen2019deep}, and E2E trains a Transformer on synthetic
equation-data pairs \citep{kamienny2022end}.
These mechanisms can exploit numerical feedback, but their candidate-generation
behavior is shaped by the chosen search rules or SR-specific learned priors.
This can limit generation flexibility across targets with different operator
combinations, nested functions, and variable interactions.

Second (\emph{Underutilized Search Discoveries}), most LLM-based
methods fail to reliably recover ground-truth expressions because they fall
short in effectively leveraging the LLM itself to thoroughly digest past
expressions or adaptively guide future search. Although many LLM-based methods reuse previously discovered expressions
and structures together with their evaluation scores, these search
discoveries are typically incorporated through
prompts, memories, libraries, or search operators while the underlying LLM
remains fixed \citep{merler2024context,shojaee2025llm,grayeli2024symbolic}.
As a result, past discoveries can influence subsequent search, but are not
progressively internalized by the LLM itself to reshape future candidate
generation.

To address these limitations, we propose \methodname\
(\underline{Di}scovery-\underline{G}uid\underline{e}d LLM Tuning for
\underline{S}ymbolic Regression via Gene\underline{t}ic Programming), which turns discoveries from genetic programming (GP) into online training signals for an LLM-derived symbolic policy. To render greater flexibility in generation, \methodname\ repurposes a pretrained decoder block from a lightweight LLM into a symbolic policy for GP population seeding.
This LLM-derived policy maps expression-tree context
to distributions over symbolic tokens. By leveraging the broader modeling capacity of the LLM, the policy supports more flexible candidate generation across diverse expression structures. The generated
expressions initialize GP populations for multi-generation evolutionary
refinement. To more thoroughly utilize search discoveries, high-quality expressions
discovered by GP are fed back to adapt the LLM policy online.
The adapted policy then seeds the next GP search. Through this online adaptation, the policy progressively internalizes past
search discoveries and translates them into improved
subsequent population generation. 

Experiments on Nguyen and Livermore show that \methodname\ outperforms the
strongest competitor by up to $2.59\%$ in recovery.
Among the evaluated LLM-based baselines, it improves recovery by at least
$54.91\%$ while using at most $17.4\%$ of the mean search time of the strongest
competitor. On 122 SRBench BlackBox datasets, \methodname\ achieves the highest
cross-dataset median test $R^2$ among the compared methods while
maintaining moderate model complexity. Our main contributions are summarized as follows: 

\begin{itemize}
    \item \textbf{LLM-derived symbolic policy.}
    We repurpose a pretrained decoder block from a lightweight LLM
    into a symbolic policy that maps expression-tree context to distributions
    over symbolic tokens for GP population seeding.
    By leveraging the broader modeling capacity of the LLM, this design
    supports more flexible candidate generation across diverse expression
    structures.

    \item \textbf{Online adaptation from search discoveries.}
    We feed high-quality expressions discovered by GP back to update the LLM
    policy online.
    This allows the LLM to progressively internalize past search discoveries
    and use them to improve subsequent population
    generation.

    \item \textbf{Evaluation.}
    Experiments on two ground-truth benchmarks and 122
    SRBench BlackBox datasets demonstrate strong recovery and predictive
    performance (Figure~\ref{fig:blackbox_teaser}).
\end{itemize}

%% file: sections/2-prelim.tex
\section{Preliminaries}
\label{sec:preliminaries}

\noindent\textbf{Notation.}
We use calligraphic capitals for sets and collections, bold lowercase letters
for vectors, and bold uppercase letters for matrices. Table~\ref{tab:main_symbols}
summarizes the main notation.

\begin{table}[t]
\centering
\footnotesize
\caption{Main notation.}
\label{tab:main_symbols}
\begin{tabularx}{\linewidth}{@{}lX@{}}
\toprule
Symbol & Meaning \\
\midrule
$\mathcal{D},\mathcal{A}$ & Observations and symbolic vocabulary. \\
$\phi,\phi_t,L$ & Expression sequence, its token at step $t$, and length. \\
$p_t,s_t$ & Parent and sibling tokens before generation step $t$. \\
$\pi_{\theta},F_{\theta}$ & Symbolic policy and pretrained decoder block. \\
$\mathbf{E}_p,\mathbf{E}_s$ & Parent and sibling embedding tables. \\
$\mathbf{W}_o,\mathbf{b}_o$ & Symbolic output projection and bias. \\
$R(\phi),\mathbf{m}_t$ & Training reward and generation-constraint vector. \\
$N,S,M$ & Population size, GP generations per restart, and elite limit. \\
$\mathcal{B},\mathcal{E},\mathcal{C}$ & Generated batch, GP elites, and combined feedback batch. \\
$\mathcal{C}_{\epsilon},\mathcal{Q}$ & High-reward feedback batch and persistent priority queue. \\
$K,\epsilon$ & Queue capacity and reward-tail parameter. \\
$B,b$ & Nominal evaluation budget and accumulated evaluation count. \\
\bottomrule
\end{tabularx}
\end{table}

\paragraph{Symbolic regression and expression trees.}
Given training observations $\mathcal{D}=\{(\mathbf{x}_i,y_i)\}_{i=1}^{n}$,
with $\mathbf{x}_i\in\mathbb{R}^{d}$, symbolic regression (SR) searches for
an expression whose predictions fit the observed relationship. Let
$\mathcal{A}$ be a predefined vocabulary of operators, input variables,
and optional free-constant tokens. Free constants are coefficients fitted
to the observations. No task-specific scientific descriptions, variable
semantics, or physical units are supplied.
An expression tree has operators at internal nodes and variables or constants
at leaves. An operator's \emph{arity} is its number of inputs: addition has
two inputs and sine has one. We represent an expression by its pre-order
sequence $\phi=(\phi_1,\ldots,\phi_L)$, visiting each node before its
children, from left to right.
Each $\phi_t\in\mathcal{A}$ labels one node, and $L=|\phi|$ is the number
of nodes. A complete sequence and the known arities determine the tree.
For example, $\sin(x_1)+x_2$ corresponds to
$\phi=(+,\sin,x_1,x_2)$. We write $f_{\phi}$ for the resulting function,
after fitting free constants when present. Its training-data reward
$R(\phi)$ measures fit, with larger values indicating better predictions.

\paragraph{Genetic programming.}
Genetic programming (GP) searches over expression populations using
fitness-based selection, subtree crossover, and mutation
\citep{koza1994genetic}. These operations select promising candidates,
recombine subexpressions, and introduce structural variations, respectively.

%% file: sections/3-method.tex
\section{Method}
\label{sec:method}
\label{sec:method_overview}

We present \methodname\ through three stages: \emph{expression generation}
with an LLM-derived symbolic policy (Section~\ref{sec:method_generate}),
\emph{candidate refinement} through GP (Section~\ref{sec:method_refine}),
and \emph{adaptation from search discoveries} to improve subsequent population
generation (Section~\ref{sec:method_adapt}). These stages repeat across
GP restarts. Figure~\ref{fig:method_overview} illustrates the framework,
and Algorithm~\ref{alg:digest} summarizes the procedure.

\input{figures/overview}

\subsection{Expression Generation}
\label{sec:method_generate}
\label{sec:method_policy}
\label{sec:method_generator}

To render greater flexibility in generation, \methodname\ repurposes a
pretrained decoder block from a lightweight LLM into a symbolic policy
$\pi_\theta$ for GP population seeding, where $\theta$ denotes its trainable parameters. Compared with conventional search mechanisms or SR-specific generators,
the policy leverages the broader modeling capacity of LLM
to model diverse expression structures.

\paragraph{Expression-tree context.}
Symbolic expressions are naturally represented as trees, whereas autoregressive LLMs operate on sequences. We therefore encode the evolving expression tree as a sequence of local structural contexts along its pre-order traversal. This converts tree construction into a sequential prediction problem that can
be naturally modeled by an autoregressive LLM. Before generating 
the next symbolic token $\phi_t$, the generated prefix $\phi_{<t}$ determines its parent $p_t$, the operator to which the next node attaches. The generated prefix $\phi_{<t}$ also determines its sibling $s_t$, which is the root of the previously completed sibling subtree.
We use $\mathrm{EMPTY}$ when the parent or sibling context is absent. For the running example $\sin(x_1)+x_2$, Table~\ref{tab:expr_tree_context} shows the structural context at each generation step, and Figure~\ref{fig:expr_tree_context} shows the
corresponding expression tree.

\noindent
\begin{minipage}[t]{0.7\linewidth}
\vspace{0pt}
\centering
\captionsetup{hypcap=false,skip=2pt}
\captionof{table}{Parent and sibling context for $\sin(x_1)+x_2$.}
\label{tab:expr_tree_context}

\scriptsize
\setlength{\tabcolsep}{3.5pt}
\renewcommand{\arraystretch}{0.95}
\begin{tabular}{@{}lcccc@{}}
\toprule
Generation step $t$ & 1 & 2 & 3 & 4 \\
\midrule
Next token $\phi_t$ & $+$ & $\sin$ & $x_1$ & $x_2$ \\
Parent $p_t$ & $\mathrm{EMPTY}$ & $+$ & $\sin$ & $+$ \\
Sibling $s_t$ & $\mathrm{EMPTY}$ & $\mathrm{EMPTY}$ & $\mathrm{EMPTY}$ & $\sin$ \\
\bottomrule
\end{tabular}
\end{minipage}
\hfill
\begin{minipage}[t]{0.25\linewidth}
\vspace{0pt}
\centering
\includegraphics[width=0.46\linewidth]{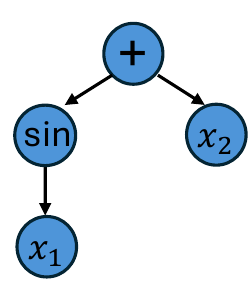}
\captionsetup{hypcap=false,skip=2pt}
\captionof{figure}{Expression tree for $\sin(x_1)+x_2$.}
\label{fig:expr_tree_context}
\end{minipage}

At the first step, both the parent and sibling contexts are $\mathrm{EMPTY}$. Before generating token $x_2$, the parent is $+$ and the sibling is $\sin$, which is the root of the completed subtree $\sin(x_1)$
(Figure~\ref{fig:expr_tree_context}).
Thus, the structural context can be constructed incrementally from the
generated prefix.

\paragraph{Pretrained symbolic interface.}
The structural contexts above are defined over the symbolic vocabulary
$\mathcal{A}$, whereas the pretrained decoder is learned over text tokens.
We bridge the two spaces by initializing a symbolic input/output interface
from pretrained token representations.
For each symbol $a\in\mathcal{A}$, we assign a fixed textual label $c(a)$,
such as $c(\texttt{mul})=\texttt{multiply}$.
If the tokenizer maps this label to token IDs
$z_{a,1},\ldots,z_{a,m_a}$, its symbolic representation is initialized as
$\mathbf{e}_a=\frac{1}{m_a}\sum_{j=1}^{m_a}
\mathbf{E}_{\mathrm{LM}}[z_{a,j}]$,
where $\mathbf{E}_{\mathrm{LM}}$ is the pretrained token-embedding table.
These representations initialize the parent and sibling embeddings
$\mathbf{E}_p,\mathbf{E}_s$ and the symbolic output projection
$\mathbf{W}_o$.
The resulting components are separately parameterized and jointly optimized
with the pretrained decoder block.
Appendix~\ref{app:symbolic_interface} provides the complete initialization
details.

\paragraph{From structural context to symbolic prediction.}
The parent-sibling contexts are interleaved into a structural sequence
\begin{equation}
    \mathbf{U}_t=
    [\mathbf{E}_p[p_1],\mathbf{E}_s[s_1],\ldots,
     \mathbf{E}_p[p_t],\mathbf{E}_s[s_t]].
    \label{eq:structural_inputs}
\end{equation}
The pretrained decoder block $F_\theta$ transforms this sequence.
Because each generation step contributes a parent and a sibling position,
we use the hidden state at the current sibling position,
$\mathbf{h}_{2t}=[F_\theta(\mathbf{U}_t)]_{2t}$, to predict the next symbol.
For the running example, after generating the prefix
$(+,\sin,x_1)$, the context $(p_4,s_4)=(+,\sin)$ is appended before
predicting token $x_2$.

The resulting symbolic policy is
\begin{equation}
    \pi_{\theta}(\phi_t=a\mid\phi_{<t})=
    \left[\operatorname{softmax}\!\left(
        \mathbf{W}_o\operatorname{LN}(\mathbf{h}_{2t})
        +\mathbf{b}_o+\mathbf{m}_t
    \right)\right]_a,
    \label{eq:symbolic_prediction}
\end{equation}
where $\operatorname{LN}$ is the trainable output LayerNorm,
$\mathbf{b}_o$ is the symbolic output bias, and $\mathbf{m}_t$ is the additive mask for the generation constraints.
Repeating this autoregressive prediction constructs a complete expression.
At each restart, \methodname\ samples $N$ expressions
$\mathcal{B}=(\phi^{(1)},\ldots,\phi^{(N)})$, which form the starting
population for GP refinement.

\subsection{Candidate Refinement}
\label{sec:method_refine}
\label{sec:method_search_learning}

Generated expressions provide starting hypotheses, but language pretraining
alone does not optimize them for the observed data.
We therefore place GP between generation and adaptation as a
numerically guided refinement stage.
Inspired by \citet{mundhenk2021symbolic}, GP starts from the
policy-generated population and explores structural variations under
task-specific numerical fitness.
Rather than asking the LLM to supply every structural improvement through
generation, this stage uses multi-generation evolutionary search to turn
numerical feedback into high-quality search discoveries.
These discoveries provide training signals for the policy adaptation
described in Section~\ref{sec:method_adapt}.

\paragraph{GP refinement.}
At each restart, the generated batch $\mathcal{B}$ initializes a fresh GP
population. \textsc{RefineGP} in Algorithm~\ref{alg:digest} evolves it through
$S$ generations of tournament selection, subtree crossover, and mutation.
During this search, numerically valid candidates receive the training reward
$R(\phi)=1/(1+\operatorname{NRMSE}(\phi))$, where NRMSE is the normalized
root-mean-square error on the training observations
(Appendix~\ref{app:search_details}).
A hall-of-fame archive records high-reward expressions encountered during
the search. The routine returns up to $M$ archived expressions as an elite
batch $\mathcal{E}$, together with the GP evaluation count $n_{\mathrm{GP}}$.

\paragraph{Search discoveries.}
After refinement, we form the feedback batch
$\mathcal{C}=\operatorname{Concat}(\mathcal{B},\mathcal{E})$
and obtain its training rewards.
Thus, $\mathcal{C}$ contains both expressions proposed by the policy and
high-quality expressions discovered through GP refinement, providing the
search discoveries used for online adaptation in the next stage.

\subsection{Adaptation from Search Discoveries}
\label{sec:method_adapt}
\label{sec:method_feedback}

To more thoroughly utilize search discoveries, we turn high-reward expressions
into training targets for the LLM-derived policy. Rather than reusing past discoveries only as inference-time context for a
fixed LLM, as in most existing LLM-based SR
methods such as ~\citet{merler2024context,shojaee2025llm}, we incorporate them through
parameter updates that shape subsequent generation.
\paragraph{Retaining search discoveries.}
A persistent priority queue $\mathcal{Q}$ keeps up to $K$ of the
highest-reward distinct expressions submitted across restarts.
For the current feedback batch $\mathcal{C}$, we form the high-reward batch
$\mathcal{C}_{\epsilon}=
[\phi\in\mathcal{C}:R(\phi)\geq q_{1-\epsilon}]$,
where $q_{1-\epsilon}$ is the empirical $(1-\epsilon)$ quantile of its
rewards. Only the highest-reward member of $\mathcal{C}_{\epsilon}$
is submitted to $\mathcal{Q}$.

\paragraph{Online policy adaptation.}
We adapt the LLM-derived policy toward retained high-reward expressions
using priority queue training (PQT) \citep{abolafia2018neural}, with
auxiliary entropy regularization to encourage exploration.
For one expression $\widetilde{\phi}$ sampled uniformly from $\mathcal{Q}$,
we minimize
\begin{equation}
    \mathcal{L}(\theta)=
    -w_{\mathrm{PQT}}\log\pi_{\theta}(\widetilde{\phi})
    -\frac{\lambda_H}{|\mathcal{C}_{\epsilon}|}
        \sum_{\phi\in\mathcal{C}_{\epsilon}}H_{\theta}(\phi).
    \label{eq:pqt_objective}
\end{equation}
Here the expression log-likelihood is
$\log\pi_{\theta}(\phi)=
\sum_{t=1}^{|\phi|}\log\pi_{\theta}(\phi_t\mid\phi_{<t})$,
and $H_{\theta}(\phi)$ sums conditional token entropies along the
expression's valid positions.
The weights $w_{\mathrm{PQT}}$ and $\lambda_H$ balance likelihood training
on the queue sample and entropy regularization over the current
$\mathcal{C}_{\epsilon}$.
Queue and loss details are provided in
Appendix~\ref{app:feedback_details}.

At each restart, one AdamW step jointly updates the pretrained decoder
block and symbolic interface (Appendix~\ref{app:run_settings}). The adapted policy seeds the next GP run, while the
persistent queue retains earlier discoveries for subsequent updates.

\begin{algorithm}[h]
\caption{\methodname: expression generation, candidate refinement,
and adaptation from search discoveries}
\label{alg:digest}
\label{alg:qwen_nggp}
\small
\begin{algorithmic}[1]
\Require Data $\mathcal{D}$; vocabulary $\mathcal{A}$; generation constraints,
GP settings, and stopping criterion
\Require Pretrained token embeddings and a  decoder block;
$N,S,M,K,\epsilon$, and nominal budget $B$
\Ensure Best feedback expression $\phi^{\star}$
\State Initialize $\pi_{\theta}$, the symbolic interface, and AdamW
\State $\mathcal{Q}\gets\varnothing$; $\phi^{\star}\gets\mathrm{None}$;
$b\gets0$; clear the expression cache
\For{$r=1,\ldots,\lfloor B/N\rfloor$}
    \Statex \textit{Expression generation}
    \State Sample $\mathcal{B}=(\phi^{(1)},\ldots,\phi^{(N)})$ from $\pi_{\theta}$
    \Statex \textit{Candidate refinement}
    \State $(\mathcal{E},n_{\mathrm{GP}})\gets
        \Call{RefineGP}{\mathcal{B},\mathcal{D},S,M}$
    \State $\mathcal{C}\gets\operatorname{Concat}(\mathcal{B},\mathcal{E})$
    \State Obtain training rewards for $\mathcal{C}$, reusing cached evaluations
    \State Update $\phi^{\star}$ using rewards in $\mathcal{C}$;
        $b\gets b+N+n_{\mathrm{GP}}$
    \Statex \textit{Adaptation from search discoveries}
    \State $\mathcal{C}_{\epsilon}\gets
    [\phi\in\mathcal{C}:R(\phi)\geq q_{1-\epsilon}]$
    \State $\widehat{\phi}\gets
        \arg\max_{\phi\in\mathcal{C}_{\epsilon}}R(\phi)$
    \State Submit $\widehat{\phi}$ to $\mathcal{Q}$; keep up to $K$
        highest-reward distinct submitted expressions
    \State Sample one $\widetilde{\phi}$ uniformly from $\mathcal{Q}$
    \State Apply one AdamW update using Eq.~(\ref{eq:pqt_objective})
    \If{$b>B$ or the configured stopping condition holds}
        \State \textbf{break}
    \EndIf
\EndFor
\State \Return $\phi^{\star}$
\end{algorithmic}
\end{algorithm}

%% file: figures/overview.tex
\begin{figure}[h]
    \centering
    \includegraphics[width=1\linewidth]{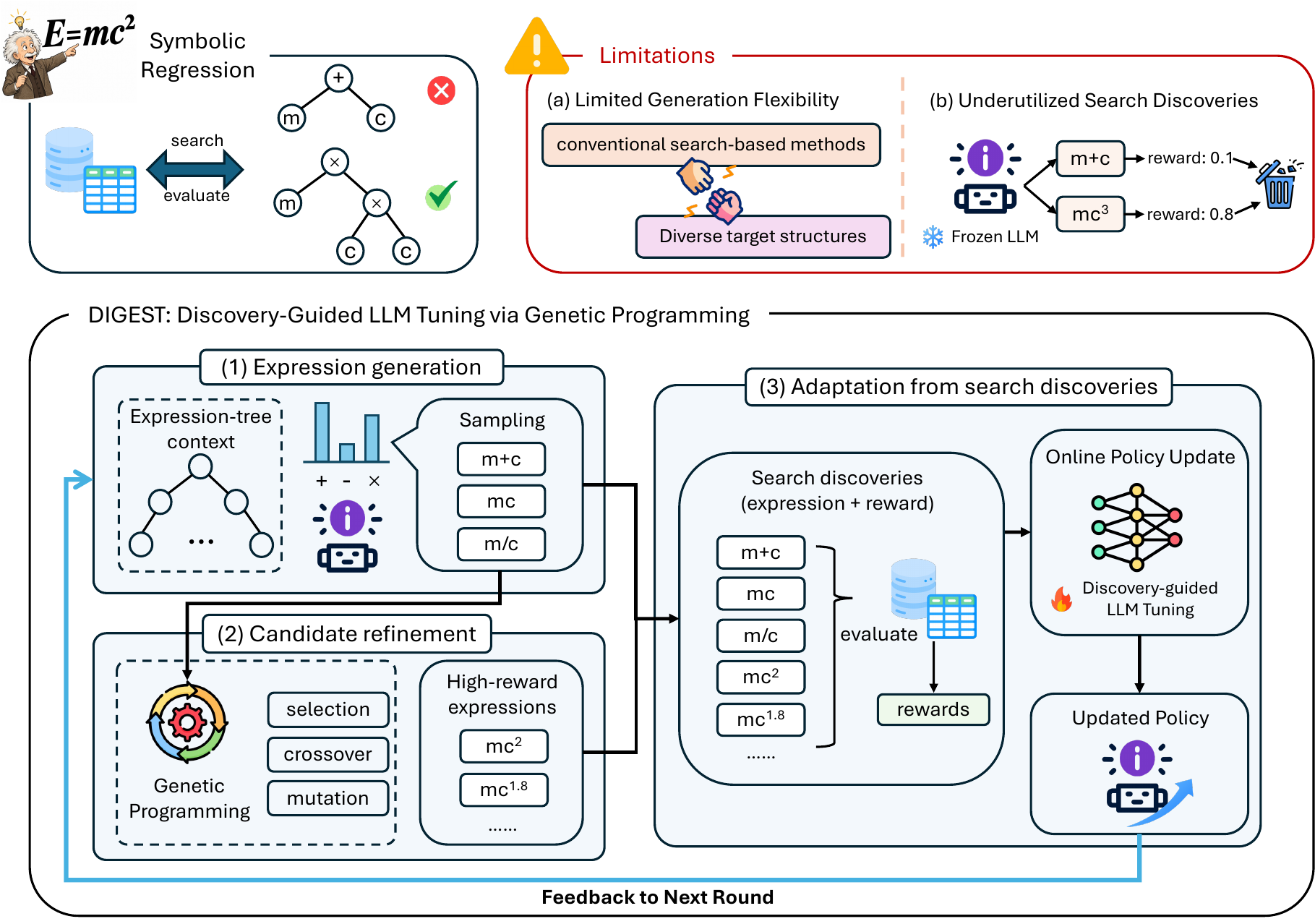}
\caption{Overview of \methodname.
    \textbf{Expression generation:} an LLM-derived symbolic policy samples a
    population from expression-tree context.
    \textbf{Candidate refinement:} GP searches for higher-reward expressions
    through evolution.
    \textbf{Adaptation from search discoveries:} policy-generated expressions and
    GP elites form a feedback batch, from which high-reward expressions
    provide training signals for online policy adaptation.
    The adapted policy seeds the next GP restart.}
    \label{fig:method_overview}
    \vspace{-15pt}
\end{figure}

%% file: sections/4-exp.tex
\section{Experiments}
\label{sec:results}

\subsection{Experimental Setup}
\label{sec:experimental_setup}

\paragraph{Datasets and metrics.}
We evaluate \methodname\ on the Nguyen \citep{petersen2019deep} and
Livermore \citep{mundhenk2021symbolic} ground-truth datasets, containing
12 and 22 target expressions, respectively, and on 122 SRBench BlackBox
datasets \citep{la2021contemporary}.
For ground-truth datasets, we report recovery averaged across equations, together with median test NMSE
and mean search time for the LLM-based comparison.
For SRBench BlackBox, we follow the standard SRBench evaluation protocol
\citep{la2021contemporary} and report cross-dataset median test $R^2$
and model size.
Detailed benchmark definitions and evaluation protocols are provided in
Appendix~\ref{app:benchmark_protocols}.

\paragraph{Baselines.}
For Nguyen and Livermore, we compare with two groups:
(1) conventional SR methods, including DSR \citep{petersen2019deep},
GEGL \citep{ahn2020guiding}, NGGP \citep{mundhenk2021symbolic},
PySR \citep{cranmer2023interpretable}, and iMCTS \citep{huang2026improving};
and (2) LLM-based methods, including ICSR \citep{merler2024context},
LLM-SR \citep{shojaee2025llm}, LaSR \citep{grayeli2024symbolic},
and RESTART \citep{li2026robust}.
For SRBench BlackBox, we compare with the 21 baseline methods reported by
\citet{la2021contemporary}, together with iMCTS
\citep{huang2026improving} and SR4MDL \citep{yu2025symbolic}.
TPSR \citep{shojaee2023transformer} is evaluated separately under its
BlackBox setting.
Descriptions and implementation details of the baseline methods are provided
in Appendix~\ref{app:baseline_implementations}.

\paragraph{Implementation.}
We instantiate \methodname\ with a decoder block from Qwen2.5-0.5B
\citep{DBLP:journals/corr/abs-2412-15115}, whereas the LLM-based
baselines use 8B backbones.
For Nguyen and Livermore, our evaluation uses 25 independent random seeds
per equation.
\methodname\ and the conventional baselines are evaluated under a
two-million-evaluation budget, while the LLM-based baselines use bounded,
method-specific search budgets to keep search costs in a comparable range
to \methodname.
All experiments run on one NVIDIA A40 GPU on the NCSA Delta
cluster.
Full configurations of \methodname\ are provided in
Appendix~\ref{app:run_settings}; baseline settings and timing details are
provided in Appendix~\ref{app:baseline_implementations}.

\subsection{Results on Ground-Truth Benchmarks}
\label{sec:main_results}
\label{sec:ground_truth_results}

\paragraph{State-of-the-art recovery.}
Table~\ref{tab:main_recovery} shows that \methodname\ achieves the highest
average recovery on both Nguyen and Livermore.
It reaches $93.33\%$ on Nguyen and $74.00\%$ on Livermore, surpassing
the strongest conventional baseline, iMCTS, by $0.08$ and $2.59$
percentage points, and the strongest LLM-based baseline, LaSR, by
$63.66$ and $54.91$ percentage points, respectively.
These gains support the central design of \methodname: turning discoveries
from GP into online training signals for an LLM-derived symbolic policy.
Per-equation results are provided in
Appendix~\ref{app:per-equation-recovery}.

\begin{table}[h]
\centering
\footnotesize
\setlength{\tabcolsep}{3.1pt}
\caption{Average recovery rates (\%) on Nguyen and Livermore.
The best result in each row is bold.}
\label{tab:main_recovery}
\begin{tabular}{@{}lccccc cccc c@{}}
\toprule
& \multicolumn{5}{c}{Conventional methods}
& \multicolumn{4}{c}{LLM-based baselines} & \\
\cmidrule(lr){2-6}\cmidrule(lr){7-10}
Benchmark & DSR & GEGL & NGGP & PySR & iMCTS
& ICSR & LLM-SR & LaSR & RESTART & \methodname \\
\midrule
Nguyen & 83.58 & 86.00 & 92.33 & 74.41 & \underline{93.25}
& 14.00 & 3.33 & 29.67 & 4.00 & \textbf{93.33} \\
Livermore & 30.41 & 56.36 & 71.09 & 46.14 & \underline{71.41}
& 16.73 & 1.27 & 19.09 & 2.18 & \textbf{74.00} \\
\bottomrule
\end{tabular}
\end{table}

\begin{figure}[h]
\centering
\begin{minipage}[t]{0.495\linewidth}
\centering
\includegraphics[width=\linewidth]{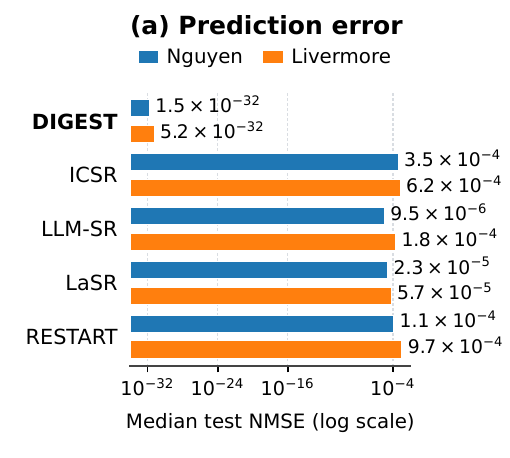}
\end{minipage}\hfill
\begin{minipage}[t]{0.495\linewidth}
\centering
\includegraphics[width=\linewidth]{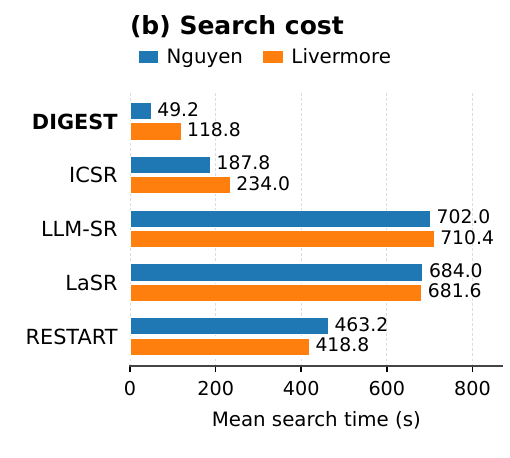}
\end{minipage}
\caption{Further comparison of \methodname\ with LLM-based baselines over
25 independent runs per equation on Nguyen and Livermore.
Left: median test NMSE on a logarithmic scale.
Right: mean search time in seconds.}
\label{fig:llm_fit_time}
\end{figure}

\paragraph{Further comparison with LLM-based methods.}
Figure~\ref{fig:llm_fit_time} further compares \methodname\ with the
LLM-based baselines in prediction error and search time.
\methodname\ achieves the lowest median test NMSE on both benchmarks,
with $1.50\times10^{-32}$ on Nguyen and $5.20\times10^{-32}$ on
Livermore, and the shortest mean search times of $49.2$ and $118.8$
seconds, respectively.
These times are only $7.2\%$ and $17.4\%$ of LaSR's mean search time.
Notably, \methodname\ repurposes a single pretrained decoder block from
Qwen2.5-0.5B, whereas all four LLM-based baselines use 8B backbones
(see Appendix~\ref{app:baseline_implementations}).

\subsection{Results on SRBench BlackBox}
\label{sec:blackbox_results}

\begin{figure}[t]
\centering
\includegraphics[width=0.68\linewidth]
{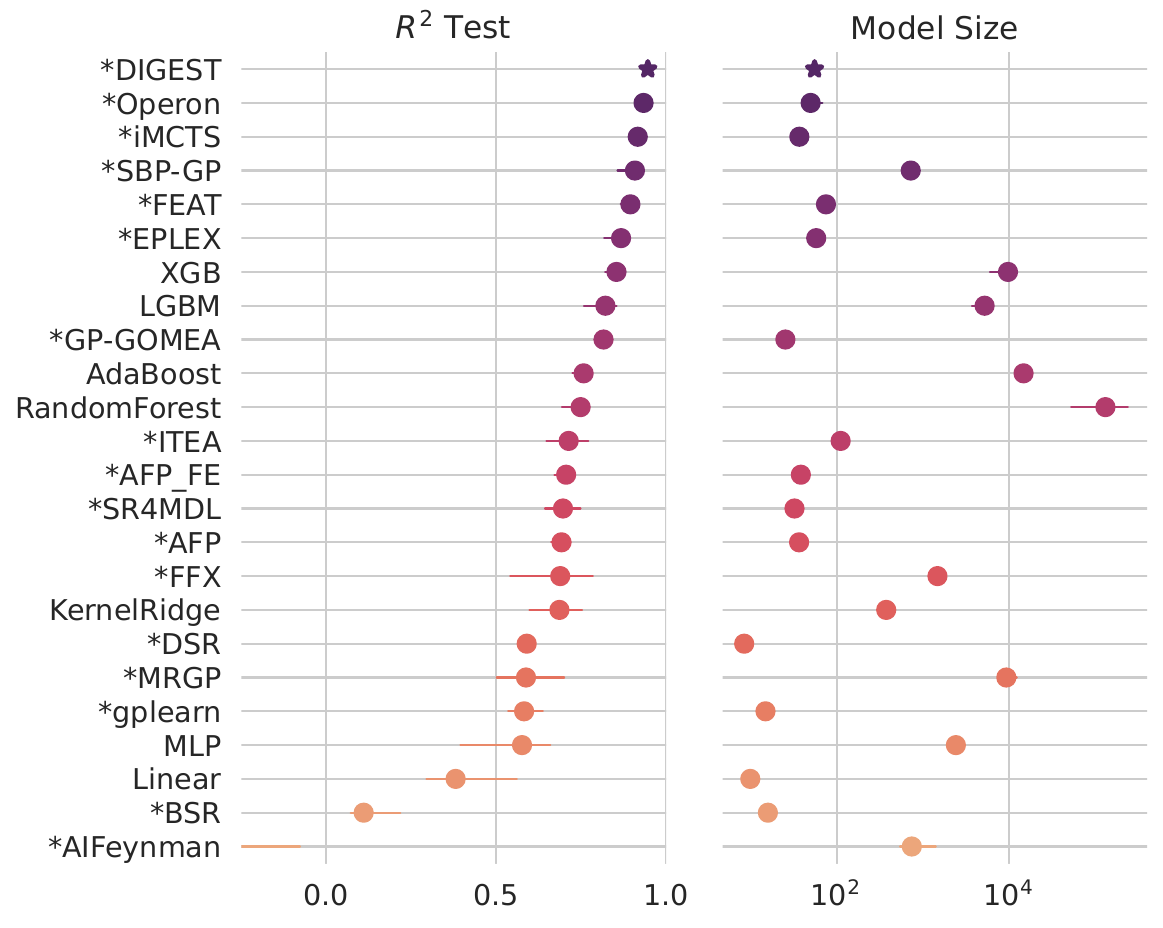}
\caption{Performance on 122 SRBench BlackBox datasets.
Left: cross-dataset median test $R^2$. Right: cross-dataset median model size.
Asterisks identify symbolic regression methods.
\methodname\ achieves the highest median predictive accuracy while
maintaining moderate model complexity.}
\label{fig:blackbox_pairgrid}
\end{figure}

On the 122 SRBench BlackBox datasets, \methodname\ achieves the highest
cross-dataset median test $R^2$ among the compared methods, including the
21 SRBench baselines, iMCTS, and SR4MDL, while maintaining moderate model
size (Figure~\ref{fig:blackbox_pairgrid}).
These results show that the benefit of LLM's learning from search discoveries
extends beyond ground-truth equation recovery to black-box regression,
where the target expression is unavailable. A Pareto view of the accuracy-complexity trade-off is provided in
Appendix~\ref{app:blackbox_additional_results} (Figure~\ref{fig:blackbox_pareto}), where \methodname\ lies on
the Pareto frontier with the highest median test $R^2$. We also compare with TPSR on the 57-dataset subset of SRBench BlackBox
used in its evaluation. \methodname\ achieves a median test $R^2$ of
$0.956$ compared with $0.945$ for TPSR, with mean model size $53.44$
compared with $95.71$.
The SRBench evaluation protocol and the separate TPSR setting are detailed in
Appendices~\ref{app:blackbox_reporting} and~\ref{app:tpsr_comparison}. 

%% file: sections/5-ablation.tex
\section{Ablation Study}
\label{sec:ablation}

We study three effects over 25 independent runs per equation (Table~\ref{tab:ablation}), with all other
evaluation settings following Section~\ref{sec:experimental_setup}.
For \textbf{online adaptation}, we compare frozen, interface-only, and
decoder-updated policies.
For \textbf{decoder depth}, we compare \methodname's single-block policy
with full-depth policies that execute all decoder blocks.
For \textbf{population seeding}, we replace the LLM-derived policy with
uniform random seeding while retaining the same GP refinement.
Except for the targeted change in each ablation, all other components and
settings of \methodname\ are kept unchanged.
For the full-depth controls, LoRA keeps the pretrained decoder weights
frozen and learns low-rank updates across all decoder blocks, whereas
full-dense tuning updates all decoder weights.
Detailed configurations are provided in
Appendices~\ref{app:run_settings} and~\ref{app:adaptation_controls}.

\begin{table}[h]
\centering
\footnotesize
\renewcommand{\arraystretch}{0.90}
\setlength{\tabcolsep}{4.5pt}
\caption{Ablations of online adaptation, decoder depth, and population seeding.
Recovery (\%) and mean search time (s) are reported.
Bold indicates the highest recovery within each neural policy-depth group.}
\label{tab:ablation}
\begin{tabular}{@{}lccccc@{}}
\toprule
& & \multicolumn{2}{c}{Livermore} & \multicolumn{2}{c}{Nguyen} \\
\cmidrule(lr){3-4}\cmidrule(lr){5-6}
Configuration & Trainable params
& Recovery $\uparrow$ & Time (s) $\downarrow$
& Recovery $\uparrow$ & Time (s) $\downarrow$ \\
\midrule
\multicolumn{6}{l}{\textit{Non-LLM population seeding}} \\
Random-seed GP & 0 & 66.91 & 120.0 & 91.00 & 46.8 \\
\addlinespace
\multicolumn{6}{l}{\textit{Single-block policy}} \\
Fully frozen & 0 & 72.91 & 130.8 & 90.67 & 55.2 \\
Interface-only updates & 0.027M & 72.18 & 130.2 & 90.67 & 54.6 \\
\methodname\ & 14.94M
& \textbf{74.00} & 118.8 & \textbf{93.33} & 49.2 \\
\addlinespace
\multicolumn{6}{l}{\textit{Full-depth policy}} \\
Interface-only updates & 0.027M
& 73.09 & 603.0 & \textbf{93.67} & 184.2 \\
LoRA updates & 14.87M
& \textbf{79.27} & 1213.8 & \textbf{93.67} & 398.4 \\
Full-dense updates & 357.92M
& 76.36 & 559.8 & \textbf{93.67} & 184.8 \\
\bottomrule
\end{tabular}
\end{table}

\paragraph{Effect of online adaptation.}
\label{sec:ablation_adaptation}

On Livermore, online adaptation improves recovery at both decoder depths.
For the single-block policy, \methodname\ reaches $74.00\%$, compared with
$72.91\%$ when fully frozen and $72.18\%$ with interface-only updates.
For full-depth policies, recovery rises from $73.09\%$ with interface-only
updates to $76.36\%$ with dense updates and $79.27\%$ with LoRA.
These results show that online adaptation is effective, supporting our
central idea that allowing the LLM-derived policy to more thoroughly
internalize past search discoveries improves subsequent generation.
On Nguyen, single-block adaptation improves recovery from $90.67\%$ to
$93.33\%$, while all full-depth variants reach $93.67\%$, consistent with
the more limited headroom on this benchmark.

\paragraph{Effect of Decoder Depth}
\label{sec:ablation_depth}

Full-depth adaptation improves recovery at substantially greater search cost.
On Livermore, LoRA and dense tuning reach $79.27\%$ and $76.36\%$ recovery
in $1213.8$ and $559.8$ seconds, versus $74.00\%$ in $118.8$ seconds for
\methodname.
Notably, \methodname\ and full-depth LoRA have similar trainable parameter
counts ($14.94$M versus $14.87$M), yet very different search times.
On Nguyen, full-depth variants improve recovery by only $0.34\%$.
These results support the single-block design as a favorable
accuracy-efficiency trade-off.

\paragraph{Effect of Population Seeding}
\label{sec:ablation_seeding}

On Livermore, recovery increases from $66.91\%$ with random seeding to
$72.91\%$ with the frozen single-block policy and $74.00\%$ with
\methodname.
The $6.00\%$ gain from random to frozen-LLM seeding shows that,
even before online adaptation, the pretrained LLM-derived policy provides
a useful structural prior for population generation, supporting the
design motivation of leveraging LLM modeling capacity for more
flexible candidate generation.

%% file: sections/6-related.tex
\section{Related Work}
\label{sec:related_work}

\paragraph{Conventional symbolic regression.}
Conventional SR generates candidates through predefined search mechanisms
or specialized learned generators.
Evolutionary and GP-based methods, evolve
expression populations through fitness-based selection and variation
\citep{schmidt2009distilling,burlacu2020operon,virgolin2021improving,cranmer2023interpretable}.
Other search methods use physics-inspired decomposition
\citep{udrescu2020ai} or Monte Carlo tree search, sometimes with learned
guidance
\citep{DBLP:journals/nca/LuTZW21,sun2023symbolicphysicslearnerdiscovering,kamienny2023deep,shojaee2023transformer,DBLP:conf/iclr/XuLS24,huang2026improving}.
Specialized generators include DSR, trained with risk-seeking policy
gradients \citep{petersen2019deep}, and models such as E2E trained on
synthetic equation-data pairs
\citep{biggio2021neural,valipour2021symbolicgpt,kamienny2022end}.
Hybrid approaches integrate neural generation with GP refinement
\citep{mundhenk2021symbolic,landajuela2022unified}, while learned
description-length estimates also guide search \citep{yu2025symbolic}.

\paragraph{LLM-based symbolic regression.}
LLM-based approaches use general-purpose language models to guide
expression discovery.
ICSR iteratively refines functional forms from numerical observations and
previous candidates, with coefficients fitted externally
\citep{merler2024context}.
LLM-SR generates equation programs from scientific descriptions and reuses
high-scoring candidates in subsequent prompts \citep{shojaee2025llm},
while LaSR augments GP with LLM-guided genetic operations and an evolving
concept library \citep{grayeli2024symbolic}.
Other approaches use simulation feedback \citep{ma2024llm},
domain-guided self-evaluation \citep{song2025llm}, reflective prompting
\citep{wang2025drsr}, or reusable structure libraries \citep{li2026robust}.
These methods reuse search discoveries through inference-time context,
self-evaluation, libraries, or search mechanisms without updating LLM
parameters during search.
In contrast, \methodname\ uses discovered high-quality expressions to
update an LLM-derived symbolic policy online, shaping subsequent generation.

%% file: sections/7-con.tex
\section{Conclusion}
\label{sec:conclusion}

In this work, we propose \methodname, which turns GP search discoveries
into online training signals for an LLM-derived symbolic policy in symbolic
regression.
To render more flexible candidate generation, \methodname\ repurposes a
pretrained decoder block from a lightweight LLM into a symbolic policy that
maps expression-tree context to distributions over symbolic tokens for GP
population seeding.
To more thoroughly utilize search discoveries, high-quality expressions
discovered by GP are fed back to update the policy online, allowing it to
progressively internalize past discoveries and improve subsequent population
generation.
Experiments on two ground-truth benchmarks and 122 BlackBox datasets demonstrate strong recovery and predictive performance.

%% file: sections/app.tex
\section{Implementation Details}
\label{app:implementation}

This appendix specifies the symbolic interface, search procedure, and
reported experimental configurations. The main method uses a single-block policy; full-depth configurations are adaptation controls. AdamW uses
separate learning-rate groups for the symbolic interface and decoder
parameters, together with weight decay and gradient clipping.

\subsection{Symbolic Interface}
\label{app:symbolic_interface}

\paragraph{Canonical labels and initialization.}
Table~\ref{tab:symbolic_labels} gives the fixed token-label mapping used for
symbolic embedding initialization. Variables keep generic vocabulary names
such as \texttt{x1}; unmapped symbols keep their original names.
Tokenization adds no special tokens. For a label tokenized into $m_a$
text-token IDs $z_{a,1},\ldots,z_{a,m_a}$, we initialize its symbolic
representation by averaging the corresponding pretrained token embeddings,
$\mathbf{e}_a=\frac{1}{m_a}\sum_{j=1}^{m_a}
\mathbf{E}_{\mathrm{LM}}[z_{a,j}]$,
where $\mathbf{E}_{\mathrm{LM}}$ is the pretrained token-embedding table.
No additional vector normalization is applied.

\begin{table}[ht]
\centering
\small
\caption{Canonical labels used to initialize the symbolic interface.}
\label{tab:symbolic_labels}
\begin{tabular}{@{}llll@{}}
\toprule
Symbol & Text label & Symbol & Text label \\
\midrule
\texttt{add} & \texttt{add} & \texttt{sub} & \texttt{subtract} \\
\texttt{mul} & \texttt{multiply} & \texttt{div} & \texttt{divide} \\
\texttt{sin} & \texttt{sin} & \texttt{cos} & \texttt{cos} \\
\texttt{tan} & \texttt{tan} & \texttt{exp} & \texttt{exp} \\
\texttt{log} & \texttt{log} & \texttt{sqrt} & \texttt{sqrt} \\
\texttt{n2} & \texttt{square} & \texttt{inv} & \texttt{inverse} \\
\texttt{neg} & \texttt{negative} & \texttt{const} & \texttt{constant} \\
\bottomrule
\end{tabular}
\end{table}

\paragraph{Parent indexing and missing states.}
Let $\mathcal{A}_{\mathrm{nt}}\subseteq\mathcal{A}$ denote the nonterminal
symbols. For every symbol $a\in\mathcal{A}$, the sibling embedding is
initialized to $\mathbf{e}_a$ and the corresponding output-projection row
to $\mathbf{e}_a^{\mathsf{T}}$. Parent embeddings use the same vectors for
$a\in\mathcal{A}_{\mathrm{nt}}$.
The parent and sibling tables additionally contain an empty-parent and an
empty-sibling entry, respectively. Although both missing states are written as $\mathrm{EMPTY}$
for readability, they use different table entries and initial values:
\begin{align}
    \mathbf{E}_p[\mathrm{EMPTY}]&\leftarrow
        \frac{1}{|\mathcal{A}_{\mathrm{nt}}|}
        \sum_{a\in\mathcal{A}_{\mathrm{nt}}}\mathbf{e}_a,
        \label{eq:empty_parent}\\
    \mathbf{E}_s[\mathrm{EMPTY}]&\leftarrow
        \frac{1}{|\mathcal{A}|}\sum_{a\in\mathcal{A}}\mathbf{e}_a.
        \label{eq:empty_sibling}
\end{align}
The parent table, sibling table, and output matrix receive separate copies
of their initialization vectors. They are jointly optimized with the policy
loss but are not tied to one another. The output bias starts at zero. Output
LayerNorm is newly constructed rather than copied from the pretrained
model's final normalization.

\paragraph{Structural computation and replay.}
Parent and sibling embeddings are interleaved and passed to the decoder,
with each symbolic step contributing two input positions and prediction
performed from the corresponding sibling position.
For GP-refined expressions, parent and sibling histories together with the
generation constraints are reconstructed from their pre-order sequences
before likelihood evaluation.
Only valid expression positions contribute to the training loss.
\subsection{Feedback Filtering and Online Objective}
\label{app:feedback_details}

\paragraph{Feedback batch and reward tail.}
The generated batch $\mathcal{B}$ is concatenated with at most $M$ expressions
from the current GP run's archive to form $\mathcal{C}$. No deduplication is performed at this
stage, so repeated expressions retain their multiplicity in reward quantiles
and entropy averages.

We select candidates whose rewards are at or above the
empirical $(1-\epsilon)$ quantile, computed using the
\texttt{higher} convention.

\paragraph{Queue admission and sampling.}
Only the highest-reward member of $\mathcal{C}_{\epsilon}$ is submitted to
$\mathcal{Q}$. The queue keeps
up to $K$ high-reward distinct submitted expressions. Each update samples one queue
entry uniformly; the reported queue capacity is $K=10$.

\paragraph{Loss reduction.}
The sequence negative log-likelihood and entropy are
\begin{align}
    -\log\pi_{\theta}(\phi)
    &=-\sum_{t=1}^{|\phi|}\log\pi_{\theta}(\phi_t\mid\phi_{<t}),\\
    H_{\theta}(\phi)
    &=\sum_{t=1}^{|\phi|}
        H\!\left(\pi_{\theta}(\cdot\mid\phi_{<t})\right).
\end{align}
These expressions refer to valid positions retained by the loss mask. Negative log-likelihood uses
the sampled queue trajectory; entropy is averaged over the current filtered
feedback batch, which may contain both policy-generated and GP-refined
expressions. Neither sequence quantity is divided by its length.

For a fixed candidate batch with finite
rewards, the highest-reward member survives reward-tail filtering.
Consequently, changing $\epsilon$ does not change that batch's queue
nominee; its direct effect is on the trajectories used for entropy
regularization. This can change later updates and search outcomes.

\subsection{GP, Evaluation, and Budget Accounting}
\label{app:search_details}
\label{app:gp_settings}

\paragraph{Training reward.}
For numerically valid predictions, the training reward is
\begin{equation}
\begin{aligned}
    \operatorname{NRMSE}(\phi)
    &=\frac{1}{\sigma_y}\sqrt{\frac{1}{n}
        \sum_{i=1}^{n}\big(y_i-f_{\phi}(\mathbf{x}_i)\big)^2},\\
    R(\phi)&=\frac{1}{1+\operatorname{NRMSE}(\phi)}.
\end{aligned}
\label{eq:training_reward}
\end{equation}
Here $\sigma_y$ is the standard deviation of the training targets.
Free coefficients, when present, are fitted to the training observations
before fitness evaluation. This reward is separate from the test-set
recovery criterion.

The recovery criterion for Nguyen and Livermore is specified in
Appendix~\ref{app:ground_truth_protocol}.

\paragraph{Genetic operations.}
GP uses tournament selection and one-point subtree crossover. Mutation
chooses uniformly among subtree replacement, node replacement, insertion,
and shrink. If a decorated
genetic operation violates a configured constraint, the offspring is
replaced by a copy of a randomly selected input parent. The full generated
batch initializes each GP run. The seeded DEAP individuals initially have
unset fitness. Fitness values are assigned when offspring are evaluated,
reusing cached rewards where available; there is no separate pass that
assigns fitness to every initial seed before the first tournament.
After GP refinement, the generated batch and archive returns are concatenated
and their training rewards are obtained, reusing cached evaluations.
The GP population and per-run archive reset at each restart; policy parameters,
AdamW moment estimates, the priority queue, expression cache, and best-so-far
expression persist.

\paragraph{Budget and stopping.}
The evaluation counter increases by $N$ for policy samples and by the GP
run's \texttt{nevals}. The latter counts individuals submitted for fitness
evaluation and can include cache hits. The budget is not a count of unique
uncached expressions or all numerical calls inside coefficient fitting.
For nominal budget $B$, the loop permits at most $\lfloor B/N\rfloor$
restarts and checks $b>B$ after queue processing and the online update.
The final restart may therefore exceed $B$. Configured stopping conditions
are checked at the same restart boundary. The search loop's best-so-far
expression is updated from the combined feedback batch.

\paragraph{Runtime.}
The internal timer measures search and online adaptation. It starts after
task and controller initialization, excluding model loading and other
initialization costs.

\subsection{Main Experimental Configuration}
\label{app:run_settings}
\label{app:adaptation_details}

\paragraph{Pretrained source and decoder configuration.}
We instantiate the pretrained source with Qwen2.5-0.5B
\citep{DBLP:journals/corr/abs-2412-15115}.
The main policy executes and trains only decoder block 10
(zero-based indexing), together with the parent and sibling
embeddings, symbolic output LayerNorm, projection, and bias.
The original text embedding and final normalization are not
used. The policy has approximately 14.94M trainable parameters,
including the symbolic interface.

\paragraph{Shared settings.}
The generated population has $N=500$ expressions. Each GP run returns at
most $M=1$ archive expression. Crossover and mutation probabilities are
both $0.5$, the tournament size is five, and the maximum generated
mutation-subtree depth is three. The queue capacity is $K=10$, the
reward-tail parameter is $\epsilon=0.02$, and each restart uses one queue
sample and one AdamW step. Sampling temperature is one. Weight decay is
$0.01$, global gradient-norm clipping is $1.0$, and the symbolic-interface
learning rate is $10^{-5}$.

\paragraph{Vocabulary and constraints.}
The operator set is
$\{+,-,\times,\div,\sin,\cos,\exp,\log\}$
for all three benchmark suites.
Nguyen and Livermore use input-variable terminals without free-constant
tokens; SRBench additionally includes the free-constant placeholder
\texttt{const}.

For Nguyen and Livermore, the active hard constraints enforce lengths from
4 to 30, prohibit immediately composed inverse unary operators
($\exp$ and $\log$), and prohibit nested trigonometric operators.

For SRBench, the length range is 4 to 60 and at most 20 \texttt{const}
tokens are permitted. Inverse-unary and nested-trigonometric constraints
remain active. The \texttt{const} token cannot be the only unique child
of an operator, and every expression must contain an input variable.
The generic relational, uniform-arity, soft-length, and language-model
priors are disabled in all three settings.

\paragraph{Nguyen and Livermore.}
Both use a nominal budget of $2\times10^6$ expression evaluations,
$S=25$ GP generations per restart, and maximum expression length $30$.
The decoder learning rate is $10^{-6}$ on Nguyen and $3\times10^{-6}$ on
Livermore. PQT weights are $100$ and $200$, respectively; entropy weights
are $0.001$ and $0.005$. Neither vocabulary includes a free-constant token.
Recovery-based early stopping is enabled.

\paragraph{SRBench BlackBox.}
The nominal budget is $5\times10^5$, with $S=50$, maximum expression length
$60$, decoder learning rate $10^{-6}$, PQT weight $100$, and entropy weight
$0.005$. Expressions may contain up to $20$ free constants.

\subsection{Full-Depth Adaptation Controls}
\label{app:adaptation_controls}

These controls are ablations of decoder depth and update parameterization,
not additional stages of the main method. Both execute all 24 decoder
blocks and densely train the symbolic interface. Their original text
embedding table remains frozen and bypassed, and the original final
normalization is replaced by the new output LayerNorm.

\paragraph{Full-model LoRA.}
Base decoder weights and biases are frozen, while low-rank updates are
trained for \texttt{q\_proj}, \texttt{k\_proj}, \texttt{v\_proj},
\texttt{o\_proj}, \texttt{gate\_proj}, \texttt{up\_proj}, and
\texttt{down\_proj}. The rank is $27$, dropout is zero, and the scaling
parameter is $27$ on Nguyen and $54$ on Livermore. The scaling parameter
is not an optimizer learning rate. LoRA uses the decoder learning-rate
group, without separate attention or feed-forward learning-rate overrides.
The Nguyen configuration uses PEFT's default initializer and an empty
\texttt{alpha\_pattern}, without module-specific scaling overrides.

With the one-variable, eight-operator Nguyen vocabulary, this control
has $14{,}873{,}865$ trainable parameters including the symbolic interface.

\paragraph{Full-model dense tuning.}
All decoder blocks and the symbolic interface are densely updated.

The corresponding one-variable Nguyen configuration has $357{,}924{,}105$
trainable parameters. Here, full model refers to the complete decoder
stack, not all original text-language-model parameters. As in the main
policy, parameter counts depend on the symbolic vocabulary.

\section{Benchmarks and Evaluation Protocols}
\label{app:benchmark_protocols}
\label{app:llm-baseline-protocol}

\subsection{Ground-Truth Benchmarks}
\label{app:ground_truth_protocol}

\paragraph{Benchmark setup.}
We follow the Nguyen and Livermore benchmark definitions of
\citet{mundhenk2021symbolic}.
Our Nguyen suite contains Nguyen-1--11 and Nguyen-12*, where Nguyen-12*
uses inputs sampled from $[0,10]$ rather than $[0,1]$; Livermore contains
22 equations.
We write $U(a,b,n)$ for $n$ points sampled uniformly from $[a,b]$ for
each input variable.
For each equation, the training and test sets are sampled separately from
the prescribed domain and then held fixed across all random seeds.
For Nguyen and Livermore, our evaluations use 25 independent random seeds
per equation.
Tables~\ref{tab:nguyen_benchmark_specs} and
\ref{tab:livermore_benchmark_specs} list the target expressions and
sampling domains.

\begin{table}[h]
\centering
\small
\setlength{\tabcolsep}{5.5pt}
\renewcommand{\arraystretch}{1.05}
\caption{Target expressions and sampling domains for the Nguyen benchmark
used in our experiments.}
\label{tab:nguyen_benchmark_specs}
\begin{tabular}{@{}lll@{}}
\toprule
Problem & Target expression & Sampling domain \\
\midrule
Nguyen-1   & $x^3+x^2+x$ & $U(-1,1,20)$ \\
Nguyen-2   & $x^4+x^3+x^2+x$ & $U(-1,1,20)$ \\
Nguyen-3   & $x^5+x^4+x^3+x^2+x$ & $U(-1,1,20)$ \\
Nguyen-4   & $x^6+x^5+x^4+x^3+x^2+x$ & $U(-1,1,20)$ \\
Nguyen-5   & $\sin(x^2)\cos(x)-1$ & $U(-1,1,20)$ \\
Nguyen-6   & $\sin(x)+\sin(x+x^2)$ & $U(-1,1,20)$ \\
Nguyen-7   & $\log(x+1)+\log(x^2+1)$ & $U(0,2,20)$ \\
Nguyen-8   & $\sqrt{x}$ & $U(0,4,20)$ \\
Nguyen-9   & $\sin(x)+\sin(y^2)$ & $U(0,1,20)$ \\
Nguyen-10  & $2\sin(x)\cos(y)$ & $U(0,1,20)$ \\
Nguyen-11  & $x^y$ & $U(0,1,20)$ \\
Nguyen-12* & $x^4-x^3+\frac{1}{2}y^2-y$ & $U(0,10,20)$ \\
\bottomrule
\end{tabular}
\end{table}

\begin{table}[h]
\centering
\small
\setlength{\tabcolsep}{5.2pt}
\renewcommand{\arraystretch}{1.04}
\caption{Target expressions and sampling domains for the Livermore benchmark
used in our experiments.}
\label{tab:livermore_benchmark_specs}
\begin{tabular}{@{}lll@{}}
\toprule
Problem & Target expression & Sampling domain \\
\midrule
Livermore-1  & $\frac{1}{3}+x+\sin(x^2)$ & $U(-10,10,1000)$ \\
Livermore-2  & $\sin(x^2)\cos(x)-2$ & $U(-1,1,20)$ \\
Livermore-3  & $\sin(x^3)\cos(x^2)-1$ & $U(-1,1,20)$ \\
Livermore-4  & $\log(x+1)+\log(x^2+1)+\log(x)$ & $U(0,2,20)$ \\
Livermore-5  & $x^4-x^3+x^2-y$ & $U(0,1,20)$ \\
Livermore-6  & $4x^4+3x^3+2x^2+x$ & $U(-1,1,20)$ \\
Livermore-7  & $\sinh(x)$ & $U(-1,1,20)$ \\
Livermore-8  & $\cosh(x)$ & $U(-1,1,20)$ \\
Livermore-9  & $x^9+x^8+x^7+x^6+x^5+x^4+x^3+x^2+x$ & $U(-1,1,20)$ \\
Livermore-10 & $6\sin(x)\cos(y)$ & $U(0,1,20)$ \\
Livermore-11 & $\frac{x^2y^2}{x+y}$ & $U(-1,1,50)$ \\
Livermore-12 & $\frac{x^5}{y^3}$ & $U(-1,1,50)$ \\
Livermore-13 & $x^{1/3}$ & $U(0,4,20)$ \\
Livermore-14 & $x^3+x^2+x+\sin(x)+\sin(x^2)$ & $U(-1,1,20)$ \\
Livermore-15 & $x^{1/5}$ & $U(0,4,20)$ \\
Livermore-16 & $x^{2/5}$ & $U(0,4,20)$ \\
Livermore-17 & $4\sin(x)\cos(y)$ & $U(0,1,20)$ \\
Livermore-18 & $\sin(x^2)\cos(x)-5$ & $U(-1,1,20)$ \\
Livermore-19 & $x^5+x^4+x^2+x$ & $U(-1,1,20)$ \\
Livermore-20 & $\exp(-x^2)$ & $U(-1,1,20)$ \\
Livermore-21 & $x^8+x^7+x^6+x^5+x^4+x^3+x^2+x$ & $U(-1,1,20)$ \\
Livermore-22 & $\exp(-0.5x^2)$ & $U(-1,1,20)$ \\
\bottomrule
\end{tabular}
\end{table}

\paragraph{Recovery criterion.}
For our Nguyen and Livermore experiments, we follow the numerical
success criterion in the released NGGP/DSO implementation \citet{mundhenk2021symbolic}: a run is
counted as recovered if its normalized mean squared error on the
noiseless test set satisfies
$\mathrm{NMSE}_{\mathrm{test}} < 10^{-12}$.
Recovery rate is the percentage of runs satisfying this criterion.
We apply the same criterion to all methods evaluated in our experiments.
For aggregate benchmark results, we average the recovery rates across
equations.
For the LLM-based comparison, we additionally report median test NMSE and
mean search time.
Timing details are provided in
Appendix~\ref{app:computation_timing}.

\subsection{SRBench BlackBox}
\label{app:blackbox_reporting}

We evaluate on the 122 SRBench BlackBox datasets
\citep{la2021contemporary}.
Each run uses a seed-specific 75/25 train/test split, without input or
target scaling.
We perform 10 independent runs per dataset with a
$5\times10^5$ evaluation budget per run and report cross-dataset median
test $R^2$ and median model size. A Pareto view of the accuracy-complexity trade-off, where \methodname\
lies on the Pareto frontier, is provided in
Appendix~\ref{app:blackbox_additional_results}.

\subsection{TPSR Evaluation Setting}
\label{app:tpsr_comparison}

TPSR \citep{shojaee2023transformer} uses an E2E Transformer pretrained
for at most ten input dimensions. Its official BlackBox setting is
described as 57 continuous-feature datasets with at most ten inputs.
Table~\ref{tab:tpsr_separate} reports our separate comparison under this
regime, using median test $R^2$ and mean model size.
These results are reported separately from the 122-task median-size analysis.

\section{Baseline Methods and Implementations}
\label{app:baseline_implementations}

\subsection{Conventional Baselines}
\label{app:conventional_baselines}

The configurations for DSR, GEGL, and NGGP follow the parameterizations
reported by \citet{mundhenk2021symbolic}. PySR and iMCTS follow the
configurations reported by \citet{huang2026improving}.

\paragraph{DSR.}
Deep Symbolic Regression uses a recurrent neural network to define a
probability distribution over mathematical expressions. A risk-seeking
policy-gradient objective trains the generator to favor high-reward
expressions \citep{petersen2019deep}.

\paragraph{GEGL.}
Genetic Expert-Guided Learning couples a neural generator with a genetic
expert that produces high-quality candidates for imitation learning
\citep{ahn2020guiding}. The symbolic-regression comparison uses the
configuration described by \citet{mundhenk2021symbolic}.

\paragraph{NGGP.}
Neural-Guided Genetic Programming combines a neural expression generator with GP searches. Expressions obtained from generation and GP
refinement provide feedback for updating the generator \citep{mundhenk2021symbolic}.

\paragraph{PySR.}
PySR evolves expression populations through fitness-based selection and
genetic variation \citep{cranmer2023interpretable}.

\paragraph{iMCTS.}
iMCTS augments Monte Carlo tree search with an extreme-bandit allocation
strategy and evolution-inspired state jumps, including mutation and
crossover, to explore nonlocal changes to candidate expressions
\citep{huang2026improving}.

\paragraph{SRBench reference methods.}
The BlackBox comparison includes the 21 baseline methods reported by
\citet{la2021contemporary}. These comprise symbolic regressors such as
Operon, FEAT \citep{DBLP:conf/iclr/CavaSTSM19}, and GP-GOMEA, together with standard machine-learning
regressors such as XGBoost \citep{DBLP:conf/kdd/ChenG16} and Random Forest \citep{DBLP:journals/ml/Breiman01}.

\paragraph{SR4MDL and TPSR.}
SR4MDL guides symbolic search with learned description-length estimates
\citep{yu2025symbolic}. TPSR incorporates Monte Carlo tree search into
Transformer-based expression generation \citep{shojaee2023transformer}.
The separate TPSR evaluation setting is described in
Appendix~\ref{app:tpsr_comparison}.

\subsection{LLM-Based Baselines}
\label{app:llm_baseline_implementations}

\begingroup
\raggedright

We evaluate ICSR \citep{merler2024context}, LLM-SR
\citep{shojaee2025llm}, LaSR \citep{grayeli2024symbolic}, and RESTART
\citep{li2026robust} using their released implementations.
We retain the default configurations of the released implementations except
for the context and budget changes described below. Key search and inference
parameters are reported for reproducibility.
We make two main changes for our evaluation.
First, to match the no-scientific-context setting of Nguyen and Livermore,
we replace task-specific scientific descriptions with neutral input and
target descriptions where applicable, while retaining generic mathematical
instructions and each method's native search-feedback mechanism.
Second, to keep search costs in a comparable range to \methodname, we use
bounded method-specific search budgets: LLM-SR is capped at 100 generated
programs, LaSR uses 20 GP iterations with 20 cycles per iteration, and
RESTART is capped at 20 generated candidates; ICSR retains its default ten
refinement iterations.
Complete prompt templates are provided in
Appendix~\ref{app:active_prompts}.

\paragraph{Shared backend for LLM-SR and LaSR.}
LLM-SR and LaSR are evaluated through the public LLM-SRBench
implementation \citep{shojaee2025llmsrbench}, using its default
configurations for the two methods except for the context and budget changes
described above.
Both use Llama-3.1-8B-Instruct \citep{DBLP:journals/corr/abs-2407-21783}.

\subsubsection{ICSR}
\label{app:icsr_implementation}

ICSR uses Llama-3-8B-Instruct
\citep{DBLP:journals/corr/abs-2407-21783}.
Its released text-only prompts are generic and remain unchanged.

\paragraph{ICSR settings.}
The search uses ten refinement iterations and up to five previous functions
with their training scores.
Temperature is $1.0$, top-$p$ is $0.90$, and top-$k$ is 60.
The output limits are 512 tokens for seed generation and 256 tokens for
refinement.
Native retries, early stopping, and the reference-equation stopping check
are retained.

\paragraph{Prompt construction.}
Each prompt contains up to 40 training points selected by the released
implementation, together with previous functions and their training scores
during refinement.
Displayed coefficients are replaced by \texttt{c}.
The complete seed-generation and refinement templates are provided in
Appendix~\ref{app:icsr_prompts}.

\subsubsection{LLM-SR}
\label{app:llmsr_implementation}

\paragraph{LLM-SR settings.}
The search uses four candidates per prompt, ten islands, and up to two prior
program examples.
In our evaluation, we cap the search at 100 generated programs.
Candidate execution uses a 30-second timeout, and SciPy BFGS is used for
coefficient optimization.
Exact-zero-training-error stopping is retained.

\paragraph{Prompt construction.}
Each request combines a neutral task instruction with the program prompt.
The initial prompt uses an affine seed program, while later requests include
up to two previously discovered programs as context.
The complete one- and two-input prompt templates are provided in
Appendix~\ref{app:llmsr_prompts}.

\subsubsection{LaSR}
\label{app:lasr_implementation}

\paragraph{LaSR settings.}
The search uses ten populations of size 33 and a maximum expression size
of 30.
In our evaluation, we use 20 GP iterations with 20 cycles per iteration.
The weights for LLM-based mutation, crossover, and random generation are
$0.005$ each.
The recent-concept window is 20 and
\texttt{num\_pareto\_context=3}.
The operator set includes \texttt{tan} and \texttt{cosh}, while the
existing nesting constraints are retained.
All LLM-assisted operations and native concept extraction and evolution
remain enabled, and each run starts with a fresh concept library.

\paragraph{Prompts and context adaptation.}
To remove task-specific scientific context, we use a neutral task
description as the first suggestion for generation, mutation, and
crossover, and remove the physical-concept clause from
\texttt{extract\_idea\_system.txt}.
The remaining active prompt templates are unchanged.
The complete templates, neutral task descriptions, and operator list are
provided in Appendix~\ref{app:lasr_prompts}.

\subsubsection{RESTART}
\label{app:restart_implementation}

\paragraph{RESTART settings.}
RESTART uses Qwen3-8B with thinking disabled.
The search uses ten islands, up to two prior program examples, and four
completions per batch.
In our evaluation, we cap the search at 20 generated candidates.
Invalid generated candidates count toward this cap, while initial E2E
expressions and conditional concept requests are counted separately.
Candidate requests use \texttt{max\_tokens=30000}; concept requests use
temperature $0.7$ and \texttt{max\_tokens=5096}.

RESTART uses the released E2E model for its native residual feedback and
retains its concept-library mechanism.
The search timer and runtime boundary are described in
Appendix~\ref{app:computation_timing}.

\paragraph{Prompts and context adaptation.}
To remove task-specific scientific context, we replace target and input
descriptions with neutral names and use mathematical rather than scientific
instructions.
Only training data are used during discovery.
Candidate prompts may include previous programs, E2E feedback consisting of
the current candidate's training error and boosted expression, and selected
library concepts.
Concept-improvement requests use the improvement between previous and new
candidates to propose new mathematical concepts.
The complete prompt templates and neutral-context substitutions are provided
in Appendix~\ref{app:restart_prompts}.

\endgroup

\subsection{Computation and Timing}
\label{app:computation_timing}

All locally evaluated methods use one NVIDIA A40 GPU per run on the
NCSA Delta cluster.
Reported search time excludes initial LLM loading and is averaged over all
runs.

\paragraph{DIGEST.}
The internal search timer of \methodname\ is described in
Appendix~\ref{app:search_details}.

\paragraph{LLM-SR and LaSR.}

LLM-SR and LaSR measure time around \texttt{discover()}, including final
training-data coefficient fitting and any first-fit Julia initialization
inside the call. Model-service startup and subsequent test evaluation are
outside this interval.

\paragraph{RESTART.}

The search timer includes E2E initialization, reloads during discovery,
and final training fits, but excludes initial LLM service startup.

\section{Detailed Experimental Results and Analyses}
\label{app:additional_results}

\subsection{Per-Equation Ground-Truth Recovery}
\label{app:per-equation-recovery}

Tables~\ref{tab:recovery_nguyen_conventional}--\ref{tab:recovery_livermore_local}
report results on all 34 equations.
Benchmark definitions and evaluation protocols are provided in
Appendix~\ref{app:ground_truth_protocol}, and baseline configurations are
provided in Appendix~\ref{app:baseline_implementations}.
Frozen and interface-only controls are summarized in
Table~\ref{tab:ablation}.

\begin{table}[htbp]
\centering
\small
\setlength{\tabcolsep}{3.7pt}
\renewcommand{\arraystretch}{1.06}
\caption{Per-equation recovery (\%) on Nguyen: conventional baselines and \methodname. The best value per row is bold.}
\label{tab:recovery_nguyen_conventional}
\label{tab:appendix-recovery-nguyen}

\begin{tabular}{@{}lcccccc@{}}
\toprule
Equation & DSR & GEGL & NGGP & PySR & iMCTS & \methodname \\
\midrule
Nguyen-1 & \textbf{100} & \textbf{100} & \textbf{100} & \textbf{100} & \textbf{100} & \textbf{100.00} \\
Nguyen-2 & \textbf{100} & \textbf{100} & \textbf{100} & 98 & \textbf{100} & \textbf{100.00} \\
Nguyen-3 & \textbf{100} & \textbf{100} & \textbf{100} & 70 & \textbf{100} & \textbf{100.00} \\
Nguyen-4 & \textbf{100} & \textbf{100} & \textbf{100} & 2 & 97 & \textbf{100.00} \\
Nguyen-5 & 72 & 92 & \textbf{100} & 89 & \textbf{100} & \textbf{100.00} \\
Nguyen-6 & \textbf{100} & \textbf{100} & \textbf{100} & \textbf{100} & \textbf{100} & \textbf{100.00} \\
Nguyen-7 & 35 & 48 & 96 & 35 & \textbf{100} & 80.00 \\
Nguyen-8 & 96 & \textbf{100} & \textbf{100} & 99 & \textbf{100} & \textbf{100.00} \\
Nguyen-9 & \textbf{100} & \textbf{100} & \textbf{100} & \textbf{100} & \textbf{100} & \textbf{100.00} \\
Nguyen-10 & \textbf{100} & 92 & \textbf{100} & \textbf{100} & \textbf{100} & \textbf{100.00} \\
Nguyen-11 & \textbf{100} & \textbf{100} & \textbf{100} & \textbf{100} & \textbf{100} & \textbf{100.00} \\
Nguyen-12* & 0 & 0 & 12 & 0 & 22 & \textbf{40.00} \\
\midrule
Average & 83.58 & 86.00 & 92.33 & 74.41 & 93.25 & \textbf{93.33} \\
\bottomrule
\end{tabular}
\end{table}

\begin{table}[htbp]
\centering
\small
\setlength{\tabcolsep}{3.7pt}
\renewcommand{\arraystretch}{1.06}
\caption{Per-equation recovery (\%) on Nguyen: LLM-based baselines, Random-seed GP, \methodname, and full-depth ablations. The best value per row is bold.}
\label{tab:recovery_nguyen_local}

\begin{tabular}{@{}lcccccccc@{}}
\toprule
Equation & ICSR & LLM-SR & LaSR & RESTART & \shortstack{Random\\GP} & \methodname & \shortstack{Full\\LoRA} & \shortstack{Full\\dense} \\
\midrule
Nguyen-1 & 20.00 & 16.00 & 28.00 & 24.00 & \textbf{100.00} & \textbf{100.00} & \textbf{100.00} & \textbf{100.00} \\
Nguyen-2 & 4.00 & 0.00 & 4.00 & 8.00 & \textbf{100.00} & \textbf{100.00} & \textbf{100.00} & \textbf{100.00} \\
Nguyen-3 & 0.00 & 12.00 & 0.00 & 0.00 & \textbf{100.00} & \textbf{100.00} & \textbf{100.00} & \textbf{100.00} \\
Nguyen-4 & 0.00 & 0.00 & 0.00 & 4.00 & \textbf{100.00} & \textbf{100.00} & \textbf{100.00} & \textbf{100.00} \\
Nguyen-5 & 0.00 & 0.00 & 0.00 & 0.00 & \textbf{100.00} & \textbf{100.00} & \textbf{100.00} & \textbf{100.00} \\
Nguyen-6 & 0.00 & 0.00 & 32.00 & 0.00 & \textbf{100.00} & \textbf{100.00} & \textbf{100.00} & \textbf{100.00} \\
Nguyen-7 & 0.00 & 0.00 & 0.00 & 0.00 & 88.00 & 80.00 & \textbf{100.00} & \textbf{100.00} \\
Nguyen-8 & 48.00 & 0.00 & \textbf{100.00} & 12.00 & \textbf{100.00} & \textbf{100.00} & \textbf{100.00} & \textbf{100.00} \\
Nguyen-9 & 0.00 & 0.00 & 80.00 & 0.00 & \textbf{100.00} & \textbf{100.00} & \textbf{100.00} & \textbf{100.00} \\
Nguyen-10 & 56.00 & 0.00 & 12.00 & 0.00 & \textbf{100.00} & \textbf{100.00} & \textbf{100.00} & \textbf{100.00} \\
Nguyen-11 & 40.00 & 0.00 & \textbf{100.00} & 0.00 & \textbf{100.00} & \textbf{100.00} & \textbf{100.00} & \textbf{100.00} \\
Nguyen-12* & 0.00 & 12.00 & 0.00 & 0.00 & 4.00 & \textbf{40.00} & 24.00 & 24.00 \\
\midrule
Average & 14.00 & 3.33 & 29.67 & 4.00 & 91.00 & 93.33 & \textbf{93.67} & \textbf{93.67} \\
\bottomrule
\end{tabular}
\end{table}

\begin{table}[htbp]
\centering
\small
\setlength{\tabcolsep}{3.7pt}
\renewcommand{\arraystretch}{1.06}
\caption{Per-equation recovery (\%) on Livermore: conventional baselines and \methodname. The best value per row is bold.}
\label{tab:recovery_livermore_conventional}
\label{tab:appendix-recovery-livermore}

\begin{tabular}{@{}lcccccc@{}}
\toprule
Equation & DSR & GEGL & NGGP & PySR & iMCTS & \methodname \\
\midrule
Livermore-1 & 3 & \textbf{100} & \textbf{100} & \textbf{100} & \textbf{100} & \textbf{100.00} \\
Livermore-2 & 87 & 44 & \textbf{100} & 32 & 95 & \textbf{100.00} \\
Livermore-3 & 66 & \textbf{100} & \textbf{100} & 97 & \textbf{100} & \textbf{100.00} \\
Livermore-4 & 76 & \textbf{100} & \textbf{100} & 90 & \textbf{100} & \textbf{100.00} \\
Livermore-5 & 0 & 0 & 4 & 30 & \textbf{78} & 16.00 \\
Livermore-6 & \textbf{97} & 64 & 88 & 0 & 19 & 48.00 \\
Livermore-7 & 0 & 0 & 0 & 0 & \textbf{15} & 0.00 \\
Livermore-8 & 0 & 0 & 0 & 0 & \textbf{22} & 0.00 \\
Livermore-9 & 0 & 12 & \textbf{24} & 0 & 2 & 20.00 \\
Livermore-10 & 0 & 0 & 24 & 26 & \textbf{35} & 16.00 \\
Livermore-11 & 17 & 92 & \textbf{100} & 92 & \textbf{100} & \textbf{100.00} \\
Livermore-12 & 61 & \textbf{100} & \textbf{100} & 18 & 91 & \textbf{100.00} \\
Livermore-13 & 55 & 84 & \textbf{100} & \textbf{100} & \textbf{100} & \textbf{100.00} \\
Livermore-14 & 0 & \textbf{100} & \textbf{100} & 15 & \textbf{100} & \textbf{100.00} \\
Livermore-15 & 0 & 96 & \textbf{100} & 74 & \textbf{100} & \textbf{100.00} \\
Livermore-16 & 4 & 12 & 92 & \textbf{93} & 72 & 80.00 \\
Livermore-17 & 0 & 4 & 68 & 72 & 69 & \textbf{100.00} \\
Livermore-18 & 0 & 0 & 56 & 3 & 29 & \textbf{100.00} \\
Livermore-19 & \textbf{100} & \textbf{100} & \textbf{100} & 88 & \textbf{100} & \textbf{100.00} \\
Livermore-20 & 98 & \textbf{100} & \textbf{100} & 76 & \textbf{100} & \textbf{100.00} \\
Livermore-21 & 2 & 64 & 24 & 0 & 41 & \textbf{100.00} \\
Livermore-22 & 3 & 68 & 84 & 9 & \textbf{100} & 48.00 \\
\midrule
Average & 30.41 & 56.36 & 71.09 & 46.14 & 71.41 & \textbf{74.00} \\
\bottomrule
\end{tabular}
\end{table}

\begin{table}[htbp]
\centering
\small
\setlength{\tabcolsep}{3.7pt}
\renewcommand{\arraystretch}{1.06}
\caption{Per-equation recovery (\%) on Livermore: LLM-based baselines, Random-seed GP, \methodname, and full-depth ablations. The best value per row is bold.}
\label{tab:recovery_livermore_local}

\begin{tabular}{@{}lcccccccc@{}}
\toprule
Equation & ICSR & LLM-SR & LaSR & RESTART & \shortstack{Random\\GP} & \methodname & \shortstack{Full\\LoRA} & \shortstack{Full\\dense} \\
\midrule
Livermore-1 & 0.00 & 0.00 & 96.00 & 0.00 & \textbf{100.00} & \textbf{100.00} & \textbf{100.00} & \textbf{100.00} \\
Livermore-2 & 0.00 & 0.00 & 0.00 & 0.00 & \textbf{100.00} & \textbf{100.00} & \textbf{100.00} & \textbf{100.00} \\
Livermore-3 & 0.00 & 0.00 & 0.00 & 0.00 & 92.00 & \textbf{100.00} & \textbf{100.00} & \textbf{100.00} \\
Livermore-4 & 0.00 & 0.00 & 0.00 & 0.00 & \textbf{100.00} & \textbf{100.00} & \textbf{100.00} & \textbf{100.00} \\
Livermore-5 & 0.00 & 0.00 & 0.00 & 0.00 & 4.00 & 16.00 & \textbf{52.00} & 36.00 \\
Livermore-6 & 0.00 & 28.00 & 0.00 & 36.00 & 4.00 & 48.00 & \textbf{76.00} & 68.00 \\
Livermore-7 & \textbf{76.00} & 0.00 & 0.00 & 0.00 & 0.00 & 0.00 & 0.00 & 0.00 \\
Livermore-8 & 56.00 & 0.00 & \textbf{100.00} & 0.00 & 0.00 & 0.00 & 0.00 & 0.00 \\
Livermore-9 & 0.00 & 0.00 & 0.00 & 0.00 & 12.00 & 20.00 & \textbf{84.00} & 64.00 \\
Livermore-10 & \textbf{56.00} & 0.00 & 12.00 & 0.00 & 8.00 & 16.00 & 44.00 & 20.00 \\
Livermore-11 & 0.00 & 0.00 & 0.00 & 0.00 & \textbf{100.00} & \textbf{100.00} & \textbf{100.00} & \textbf{100.00} \\
Livermore-12 & 0.00 & 0.00 & 0.00 & 0.00 & \textbf{100.00} & \textbf{100.00} & \textbf{100.00} & \textbf{100.00} \\
Livermore-13 & 0.00 & 0.00 & 56.00 & 0.00 & \textbf{100.00} & \textbf{100.00} & \textbf{100.00} & \textbf{100.00} \\
Livermore-14 & 0.00 & 0.00 & 0.00 & 0.00 & \textbf{100.00} & \textbf{100.00} & \textbf{100.00} & \textbf{100.00} \\
Livermore-15 & 60.00 & 0.00 & 60.00 & 0.00 & \textbf{100.00} & \textbf{100.00} & \textbf{100.00} & \textbf{100.00} \\
Livermore-16 & 0.00 & 0.00 & 56.00 & 0.00 & \textbf{80.00} & \textbf{80.00} & 40.00 & 52.00 \\
Livermore-17 & 76.00 & 0.00 & 4.00 & 0.00 & 60.00 & \textbf{100.00} & 96.00 & 96.00 \\
Livermore-18 & 0.00 & 0.00 & 0.00 & 0.00 & 36.00 & \textbf{100.00} & \textbf{100.00} & \textbf{100.00} \\
Livermore-19 & 0.00 & 0.00 & 4.00 & 12.00 & \textbf{100.00} & \textbf{100.00} & \textbf{100.00} & \textbf{100.00} \\
Livermore-20 & 32.00 & 0.00 & 16.00 & 0.00 & \textbf{100.00} & \textbf{100.00} & \textbf{100.00} & \textbf{100.00} \\
Livermore-21 & 0.00 & 0.00 & 0.00 & 0.00 & 96.00 & \textbf{100.00} & \textbf{100.00} & \textbf{100.00} \\
Livermore-22 & 12.00 & 0.00 & 16.00 & 0.00 & \textbf{80.00} & 48.00 & 52.00 & 44.00 \\
\midrule
Average & 16.73 & 1.27 & 19.09 & 2.18 & 66.91 & 74.00 & \textbf{79.27} & 76.36 \\
\bottomrule
\end{tabular}
\end{table}

\clearpage

\subsection{Accuracy--Complexity Analysis on SRBench BlackBox}
\label{app:blackbox_additional_results}

\begin{figure}[htbp]
\centering
\includegraphics[width=0.88\linewidth]
{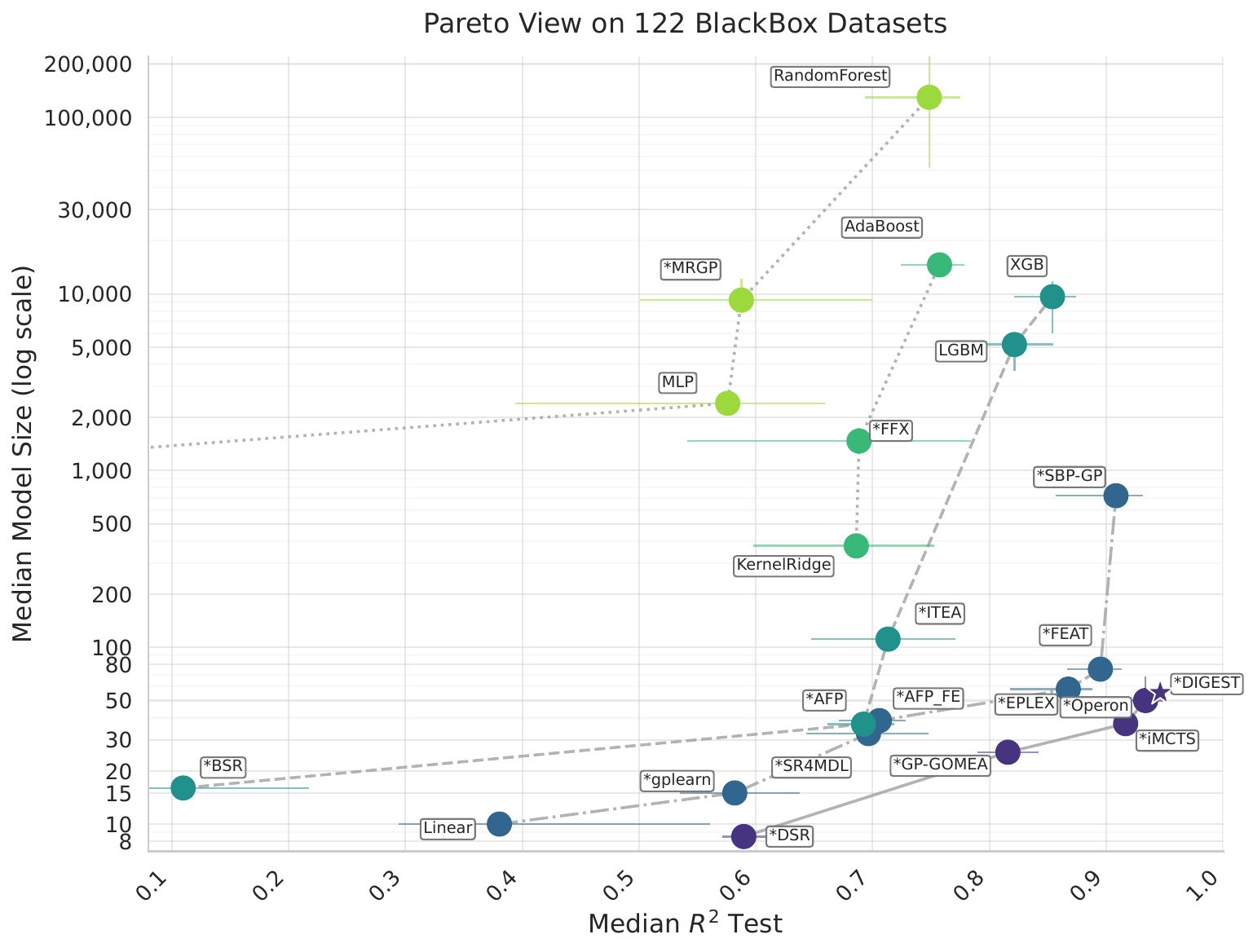}
\caption{Pareto view of cross-dataset median test $R^2$ and median model
size on 122 SRBench BlackBox datasets. \methodname\ lies on the Pareto
frontier with the highest median test $R^2$ and moderate model size.
The model-size axis is logarithmic; asterisks identify symbolic regression
methods.}
\label{fig:blackbox_pareto}
\end{figure}

\paragraph{Pareto view.}
Figure~\ref{fig:blackbox_pairgrid} reports cross-dataset median test $R^2$
and median model size separately, while Figure~\ref{fig:blackbox_pareto}
shows their joint accuracy-complexity trade-off.
\methodname\ lies on the Pareto frontier, achieving the highest median test
$R^2$ among the compared methods while maintaining moderate model size.

\paragraph{TPSR comparison.}
Table~\ref{tab:tpsr_separate} reports the separate comparison under the
setting described in Appendix~\ref{app:tpsr_comparison}, using median test
$R^2$ and mean model size.

\begin{table}[h]
\centering
\small
\caption{Comparison under the TPSR BlackBox evaluation setting.}
\label{tab:tpsr_separate}
\begin{tabular}{@{}lcc@{}}
\toprule
Method & Median test $R^2$ $\uparrow$ & Mean model size $\downarrow$ \\
\midrule
TPSR & 0.945 & 95.71 \\
\methodname & \textbf{0.956} & \textbf{53.44} \\
\bottomrule
\end{tabular}
\end{table}

\section{Active Prompt Templates}
\label{app:active_prompts}
\label{app:llmsr_lasr_prompts}
\label{app:icsr-restart-prompts}

The templates below are reproduced from the active implementations.
Dynamic expressions and feedback are filled from the training search.
Baseline implementation details are provided in
Appendix~\ref{app:llm_baseline_implementations}; benchmark definitions
and evaluation protocols are provided in
Appendix~\ref{app:benchmark_protocols}.

\begingroup
\raggedright

\subsection{ICSR Prompts}
\label{app:icsr_prompts}

Message construction and substitutions are described in
Appendix~\ref{app:icsr_implementation}.

\begin{DigestPromptBox}{Seed-generation template (complete).}
I want you to act as a mathematical function generator.
Given a set of points below, you are to come up with 5 potential functions that would fit the points. Don't worry too much about accuracy: your task is to generate a set of functions that are as diverse as possible, so that they can serve as starting points for further optimization.

To generate the functions, you will start from a set of basic operators and expressions, and combine them into something more complex. 
Your options are:
- {num_variables} independent variable symbols: {variables_list}.
- A coefficient symbol: c (there is no need to write a number - write this generic coefficient instead).
- Basic operators: +, -, *, /, ^, sqrt, exp, log, abs
- Trigonometric expressions: sin, cos, tan, sinh, cosh, tanh
- Standard constants: "pi" represents pi and "E" represents euler's constant.

Make sure there are no numbers in the functions, use the coefficient token 'c' instead.
You are required to use at least one of each variables from the available list ({variables_list}).
Analyze the points carefully: if there are any negative points in the input, sqrt and log can not be used unless the input can never be negative. Be careful about dividing by zero!
The functions should all begin with the indicators "f1 = ", "f2 = "... Only write the new function and no additional output.
Your task is to combine an arbitrary number of these basic blocks to create a complex expression. Don't be afraid to be creative and experiment! The functions should be as complex as possible, combining many different operations. Variety is key!

Points: {points}

Functions:
\end{DigestPromptBox}

\begin{DigestPromptBox}{Refinement template (complete).}
I want you to act as a mathematical function generator.
You are given a set of points with (x, y) coordinates below: 
{points}
Below are some previous functions and the error they make on the points above. The errors are arranged in order of their fit values, with the highest values coming first, and lower is better.

Your task is to give me a list of five new functions that are different from all the ones reported below, and have a lower error value than all of the ones below. Only output the new functions and nothing else.
Remember that the functions you generate should always have at most {num_variables} variables {variables_list}. 
The functions should have parametric form, using 'c' in place of any constant or coefficient. The coefficients will be optimized to fit the data. Make absolutely sure that the functions you generate are completely different from the ones already given to you.
Remember that you can combine the simple building blocks (operations, constants, variables) in any way you want to generate more complex functions. Don't be afraid to experiment!
The functions should all begin with the indicators "f1(x) = ", "f2(x) = "... 

{functions}
\end{DigestPromptBox}

\subsection{LLM-SR Prompts}
\label{app:llmsr_prompts}

Message construction is described in
Appendix~\ref{app:llmsr_implementation}.

\begin{DigestPromptBox}{Complete initial user message, one input.}
You are a helpful assistant tasked with discovering mathematical function structures from numerical data. Complete the 'equation' function below, considering the mathematical relationships among the inputs.

"""
Find the mathematical function skeleton that represents target value, given data on input variable x1.
"""

import numpy as np

#Initialize parameters
MAX_NPARAMS = 10
params = [1.0]*MAX_NPARAMS

def equation_v0(x1: np.ndarray, params: np.ndarray) -> np.ndarray:
    """ Mathematical function for target value

    Args:
        x1: A numpy array representing observations of input variable x1.
        params: Array of numeric constants or parameters to be optimized

    Return:
        A numpy array representing target value as the result of applying the mathematical function to the inputs.
    """
    output = params[0] * x1 + params[1]
    return output

def equation_v1(x1: np.ndarray, params: np.ndarray) -> np.ndarray:
    """Improved version of `equation_v0`."""
\end{DigestPromptBox}

\begin{DigestPromptBox}{Complete initial user message, two inputs.}
You are a helpful assistant tasked with discovering mathematical function structures from numerical data. Complete the 'equation' function below, considering the mathematical relationships among the inputs.

"""
Find the mathematical function skeleton that represents target value, given data on input variable x1, and input variable x2.
"""

import numpy as np

#Initialize parameters
MAX_NPARAMS = 10
params = [1.0]*MAX_NPARAMS

def equation_v0(x1: np.ndarray, x2: np.ndarray, params: np.ndarray) -> np.ndarray:
    """ Mathematical function for target value

    Args:
        x1: A numpy array representing observations of input variable x1.
        x2: A numpy array representing observations of input variable x2.
        params: Array of numeric constants or parameters to be optimized

    Return:
        A numpy array representing target value as the result of applying the mathematical function to the inputs.
    """
    output = params[0] * x1 + params[1] * x2 + params[2]
    return output

def equation_v1(x1: np.ndarray, x2: np.ndarray, params: np.ndarray) -> np.ndarray:
    """Improved version of `equation_v0`."""
\end{DigestPromptBox}

\subsection{LaSR Prompts}
\label{app:lasr_prompts}

\begin{DigestPromptBox}{Neutral task context, one input.}
The expression represents target value (y), given data on input variable x1 (x1).
\end{DigestPromptBox}

\begin{DigestPromptBox}{Neutral task context, two inputs.}
The expression represents target value (y), given data on input variable x1 (x1), and input variable x2 (x2).
\end{DigestPromptBox}

\begin{DigestPromptBox}{Available operators.}
binary operators: +, *, -, /, ^, and unary operators: exp, log, sqrt, sin, cos, tan, cosh
\end{DigestPromptBox}

\begin{DigestPromptBox}{\texttt{gen\_random\_system.txt}.}
You are a helpful assistant that proposes a mathematical expression by following three provided suggestions.
An expression must consist of the following variables: {{variables}}. All constants will be represented with the symbol C. Each expression will only use these operators: {{operators}}.
\end{DigestPromptBox}

\begin{DigestPromptBox}{\texttt{gen\_random\_user.txt}.}
Suggestion 1: {{assump1}}
Suggestion 2: {{assump2}}
Suggestion 3: {{assump3}}

Propose {{N}} expressions that would be appropriate given the suggestions. Provide short commentary for each of your decisions. End with a JSON list that enumerates the proposed expressions following this format:
```json
["expr1",
 "expr2",
 ...
 "expr{{N}}"
]
```
\end{DigestPromptBox}

\begin{DigestPromptBox}{\texttt{mutate\_system.txt}.}
You are a helpful assistant that mutates a mathematical expression by following a few provided suggestions. You will be given three suggestions and a single reference expression to mutate.
An expression must consist of the following variables: {{variables}}. All constants will be represented with the symbol C. Each expression will only use these operators: {{operators}}.
\end{DigestPromptBox}

\begin{DigestPromptBox}{\texttt{mutate\_user.txt}.}
Suggestion 1: {{assump1}}
Suggestion 2: {{assump2}}
Suggestion 3: {{assump3}}
Reference Expression: {{expr}}

Propose {{N}} expressions that would be appropriate given the suggestions and references. Provide short commentary for each of your decisions. End with a JSON list that enumerates the proposed expressions following this format:
```json
["expr1",
 "expr2",
 ...
 "expr{{N}}"
]
```
\end{DigestPromptBox}

\begin{DigestPromptBox}{\texttt{crossover\_system.txt}.}
You are a helpful assistant that recombines two mathematical expressions by following a few provided suggestions. You will be given three suggestions and two reference expressions to recombine.
An expression must consist of the following variables: {{variables}}. All constants will be represented with the symbol C. Each expression will only use these operators: {{operators}}.
\end{DigestPromptBox}

\begin{DigestPromptBox}{\texttt{crossover\_user.txt}.}
Suggestion 1: {{assump1}}
Suggestion 2: {{assump2}}
Suggestion 3: {{assump3}}
Reference Expression 1: {{expr1}}
Reference Expression 2: {{expr2}}

Propose {{N}} expressions that would be appropriate given the suggestions and references. Provide short commentary for each of your decisions. End with a JSON list that enumerates the proposed expressions following this format:
```json
["expr1",
 "expr2",
 ...
 "expr{{N}}"
]
```
\end{DigestPromptBox}

\begin{DigestPromptBox}{\texttt{extract\_idea\_system.txt}.}
You are a helpful assistant that hypothesizes about the underlying assumptions that generated a list of good and bad mathematical expressions in detailed ways. My ultimate goal is to discover what assumptions generated the observed good mathematical expressions and excludes the bad mathematical expressions. Focus more on the good expressions and their mathematical structure. Note that capital C represents an arbitrary constant.
\end{DigestPromptBox}

\begin{DigestPromptBox}{\texttt{extract\_idea\_user.txt}.}
Good Expression 1: {{gexpr1}}
Good Expression 2: {{gexpr2}}
Good Expression 3: {{gexpr3}}
Good Expression 4: {{gexpr4}}
Good Expression 5: {{gexpr5}}

Bad Expression 1: {{bexpr1}}
Bad Expression 2: {{bexpr2}}
Bad Expression 3: {{bexpr3}}
Bad Expression 4: {{bexpr4}}
Bad Expression 5: {{bexpr5}}

Propose {{N}} hypotheses that would be appropriate given the expressions. Provide short commentary for each of your decisions. Do not talk about topics related to the simplicity or complexity of the expressions. I want ideas that are unique and interesting enough to amaze the world's best mathematicians. End with a JSON list that enumerates the proposed hypotheses following this format:
```json
["hyp1",
 "hyp2",
 ...
 "hyp{{N}}"
]
```
\end{DigestPromptBox}

\begin{DigestPromptBox}{\texttt{prompt\_evol\_system.txt}.}
You are a helpful assistant that merges and refines ideas about a set of hidden mathematical expression in new, interesting, and diverse ways. My ultimate goal is to discover the underlying properties of these hidden expressions. The resulting ideas should be a nontrivial conclusion given the previous ideas.
\end{DigestPromptBox}

\begin{DigestPromptBox}{\texttt{prompt\_evol\_user.txt}.}
Idea 1: {{idea1}}
Idea 2: {{idea2}}
Idea 3: {{idea3}}
Idea 4: {{idea4}}
Idea 5: {{idea5}}

Propose {{N}} hypotheses that would be appropriate given the ideas. Provide short commentary for each of your decisions. Do not talk about topics related to the simplicity or complexity of the expressions. I want ideas that are unique and interesting enough to amaze the world's best mathematicians. End with a JSON list that enumerates the proposed hypotheses following this format:
```json
["hyp1",
 "hyp2",
 ...
 "hyp{{N}}"
]
```
\end{DigestPromptBox}

\subsection{RESTART Prompts}
\label{app:restart_prompts}

An overview of the RESTART implementation and its
no-scientific-context adaptation is provided in
Appendix~\ref{app:restart_implementation}.

\begin{DigestPromptBox}{Task descriptions.}
Find the mathematical function skeleton that represents target value, given data on input variable 1.

Find the mathematical function skeleton that represents target value, given data on input variable 1, and input variable 2.
\end{DigestPromptBox}

\begin{DigestPromptBox}{Neutral-context substitutions.}
structures for scientific systems.
  -> structures from numerical data.
scientific meaning and relationships of inputs.
  -> mathematical relationships among the inputs.
scientifically meaningful
  -> mathematically meaningful
'Harmonic Oscillator'
  -> 'Trigonometric Function'
key characteristics in scientific context
  -> key mathematical characteristics
\end{DigestPromptBox}

\begin{DigestPromptBox}{Candidate instruction.}
You are a helpful assistant tasked with discovering mathematical function structures from numerical data.                              Complete the 'equation' function below, considering the mathematical relationships among the inputs.

\end{DigestPromptBox}

\begin{DigestPromptBox}{Version docstring.}
Improved version of `equation_v{i-1}`.
\end{DigestPromptBox}

\begin{DigestPromptBox}{E2E feedback.}
    Analysis for current equation(f): 
        Given (X,y), Current equation(f) has MSE: {MSE:.6f}, 
        Definitions: current_pred = f_current(X) (the output of current equation).
        We fit a boosted equation g(X, current_pred) to approximate y.
        The learned boosted equation candidate is: {residual_eq}, 
        with MSE: {residual_mse:.6f}
\end{DigestPromptBox}

\begin{DigestPromptBox}{Next-version instruction with boosted feedback.}

    Consider previous versions of the equation with the corresponding boosted function and concepts(if any) for the problem,
        and create a new version of the equation that is more accurate.
         You can also combine terms from both the equation and boosted function from previous versions
\end{DigestPromptBox}

\begin{DigestPromptBox}{Next-version instruction without boosted feedback.}

    Consider previous versions of the equation and concepts(if any) for the problem,
        and create a new version of the equation that is more accurate.
         You can also combine terms from both the equation and function from previous versions
        while keeping the equation mathematically meaningful by retaining proper concepts.
\end{DigestPromptBox}

\begin{DigestPromptBox}{Library-concept entries.}
    Top scoring concepts for the problem:
        - {concept_name}, fit: {fit_score:.0f}/100
            definition: {definition}
            functions: {comma_separated_functions}
\end{DigestPromptBox}

\begin{DigestPromptBox}{Concept-improvement message.}
Analyze the improvement between the baseline equation and the new equation, then propose ONE NEW mathematical concept that best explains the improvement.

BASELINE EQUATION (MSE: {baseline_mse:.6f}):
{baseline_code}

NEW EQUATION (MSE: {current_mse:.6f}):
{code}

IMPROVEMENT METRICS:
- Relative improvement: {improvement_ratio:.4f} ({improvement_ratio * 100:.2f}%)
- Absolute improvement: {absolute_improvement:.6f}

Based on this improvement, identify ONE key mathematical concept that likely contributed to the better performance. Focus on:
1. New mathematical operations or functions introduced
2. Structural changes (e.g., polynomial degree, interaction terms)
3. Mathematical transformations or patterns

Provide:
- name: A concise mathematical concept name (e.g., 'Logarithmic Transform', 'Polynomial Interaction')
- definition: Brief explanation of the concept and why it improves performance
- functions: List of specific functions/operators that implement this concept

Do not output other text, explanations, comments, or natural language.
Return STRICTLY ONE JSON object following the schema below.

JSON Schema: {schema_json}
\end{DigestPromptBox}

\begin{DigestPromptBox}{Concept-output schema.}
{
  "type": "object",
  "properties": {
    "name": {
      "type": "string",
      "description": "A concise, descriptive name for the mathematical concept (e.g., 'Power Law', 'Exponential Decay', 'Trigonometric Function')"
    },
    "definition": {
      "type": "string",
      "description": "A clear mathematical definition explaining what this concept represents and its key mathematical characteristics"
    },
    "functions": {
      "type": "string",
      "description": "The mathematical functions, operations, or transformations that characterize this concept, or relevant Python code implementations"
    }
  },
  "required": [
    "name",
    "definition",
    "functions"
  ]
}
\end{DigestPromptBox}

\endgroup